\documentclass[11pt]{article}

\usepackage{acl}
\usepackage{times}
\usepackage{latexsym}
\usepackage{amsmath}
\usepackage{booktabs}
\usepackage{fix-cm}
\usepackage{multirow}
\usepackage{tikz}
\usetikzlibrary{calc}
\usepackage{array}
\usepackage{float}
\usepackage{caption}

\usepackage[T1]{fontenc}

\usepackage[utf8]{inputenc}

\usepackage{microtype}

\usepackage{inconsolata}

\usepackage{graphicx}

\title{Benchmarking Foundation Models with Retrieval-Augmented Generation in Olympic-Level Physics Problem Solving}

\author{
  \textbf{Shunfeng Zheng\textsuperscript{1}\thanks{Equal contribution.}},
  \textbf{Yudi Zhang\textsuperscript{2*}},
  \textbf{Meng Fang\textsuperscript{3}}, \\
  \textbf{Zihan Zhang\textsuperscript{1}},
  \textbf{Zhitan Wu\textsuperscript{4}},
  \textbf{Mykola Pechenizkiy\textsuperscript{2}},
  \textbf{Ling Chen\textsuperscript{1}}
\\
\\
  \textsuperscript{1}AAII, University of Technology Sydney, New South Wales, Australia \\
  \textsuperscript{2}Eindhoven University of Technology, Eindhoven, The Netherlands \\
  \textsuperscript{3}University of Liverpool, Liverpool, United Kingdom \\
  \textsuperscript{4}University of New South Wales, New South Wales, Australia
\\
{\small
\begin{tabular}{@{}c@{}}
\texttt{Shunfeng.Zheng@student.uts.edu.au}, \texttt{y.zhang5@tue.nl}, \texttt{Meng.Fang@liverpool.ac.uk} \\
\texttt{Zihan.Zhang-5@student.uts.edu.au}, \texttt{zhitan.wu@student.unsw.edu.au}, \texttt{m.pechenizkiy@tue.nl} \\
\texttt{Ling.Chen@uts.edu.au}
\end{tabular}
}\\[0.8em]
}

\begin{document}
\maketitle

\begin{abstract}
Retrieval-augmented generation (RAG) with foundation models has achieved strong performance across diverse tasks, but their capacity for expert-level reasoning—such as solving Olympiad-level physics problems—remains largely unexplored. Inspired by the way students prepare for competitions by reviewing past problems, we investigate the potential of RAG to enhance physics reasoning in foundation models.  
We introduce \textbf{PhoPile}, a high-quality multimodal dataset specifically designed for Olympiad-level physics, enabling systematic study of retrieval-based reasoning. PhoPile includes diagrams, graphs, and equations, capturing the inherently multimodal nature of physics problem solving.  
Using PhoPile, we benchmark RAG-augmented foundation models, covering both large language models (LLMs) and large multimodal models (LMMs) with multiple retrievers. Our results demonstrate that integrating retrieval with physics corpora can improve model performance, while also highlighting challenges that motivate further research in retrieval-augmented physics reasoning.  
\end{abstract}

\section{Introduction}
Physics plays a foundational role in natural sciences and engineering, underpinning progress in fields ranging from construction and aerospace to electronics and materials science~\cite{serway2000physics}. Mastering physics requires not only conceptual understanding of natural laws but also the ability to integrate them with quantitative analysis, diagrams, and symbolic reasoning.
Recent years have witnessed the exceptional performance of foundation models, including large language models (LLMs) and large multimodal models (LMMs), such as GPT-3~\cite{Brown2020LanguageMA}, 
Chat-GLM~\cite{glm2024chatglm}, GPT-4~\cite{achiam2023gpt}, GPT-3.5~\cite{openai2022chatgpt} and Gemini~\cite{Anil2023GeminiAF}. These models demonstrate strong capabilities in logic and mathematics~\cite{imani2023mathprompter, romera2024mathematical,Liu2023ImprovingLL, shi2023cooperation} and scientific domains~\cite{Singhal2023, BINGLER2022102776}, raising the prospect of AI agents that can support scientific discovery through physics reasoning. However, despite these advances, foundation models still face serious limitations: they lack domain-specific expertise~\cite{Li2023AreCA,Shen2023InCW}, frequently hallucinate factual content~\cite{ji2023survey,xiong2023can}, and struggle to consistently apply the appropriate physical principles in problem-solving.

\begin{figure*}[ht]
    \centering
    \includegraphics[width=\textwidth]{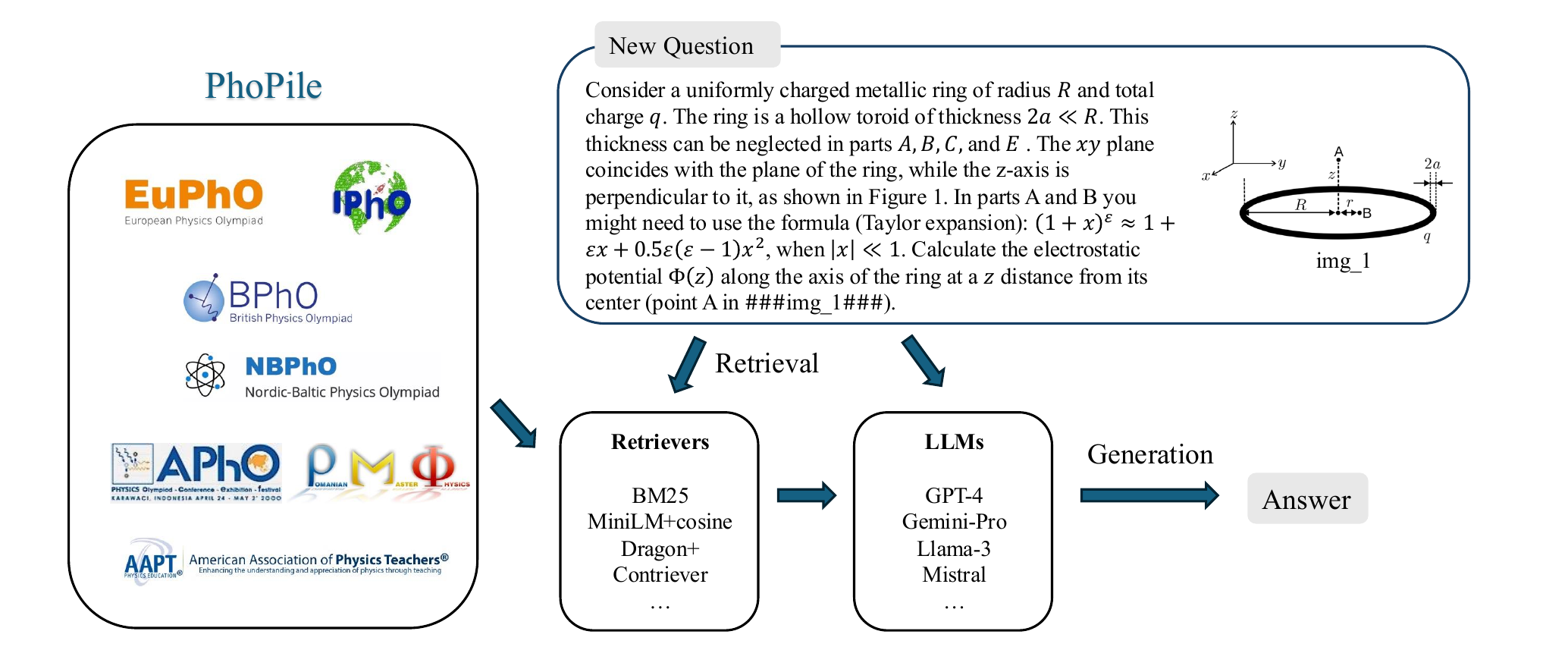}
    \caption{PhoPile and the overall workflow of foundation models with RAG.}
    \label{fig:fig1}
\end{figure*}

Retrieval-Augmented Generation (RAG) has recently emerged as a promising approach to mitigating the limitations of LLMs by integrating external knowledge sources into their workflows~\cite{caffagni2024wiki, gao2023retrieval}. 
Yet, its effectiveness for physics reasoning remains largely unexplored. 
In this context, RAG offers a natural solution: just as a student preparing for an exam consults past competition problems to recall formulas and problem-solving strategies, an LLM equipped with retrieval can access relevant laws, examples, and reasoning patterns from a curated retrieval corpus. 
This mechanism not only improves factual accuracy but also helps guide the selection and application of physical principles in diverse contexts.  

A key obstacle, however, is the absence of suitable benchmarks. 
Whereas mathematics has benefited from a rich ecosystem of high-quality datasets and benchmarks~\cite{zheng2021minif2f, hendrycks2021measuring, cobbe2021gsm8k, clark2018think, azerbayev2024llemmaopenlanguagemodel,fang2024mathodysseybenchmarkingmathematicalproblemsolving}, physics-specific evaluations are scarce. 
Existing natural science datasets~\cite{welbl2017crowdsourcingmultiplechoicescience,lu2022learn,chen2023theoremqa} contain only a small number of low-difficulty, text-only physics problems. 
OlympiadBench~\cite{he2024olympiadbench} raises the level of challenge but evaluates models in isolation, without retrieval. 
Furthermore, evaluation itself poses unique challenges: physics answers may take diverse forms—numerical, symbolic, or diagrammatic—making automatic grading substantially more difficult than in mathematics.



To address these gaps, we introduce \textbf{PhoPile}\footnote{Data and code available at: \url{https://github.com/aialt/PhoPile}}, the first multimodal RAG benchmark for physics Olympiad problems, as illustrated in Figure~\ref{fig:fig1}. 
PhoPile consists of 390 Olympiad-level physics problems from 2019–2021 for evaluation, along with 2,662 problems from earlier years that serve as an external retrieval database.  
Using PhoPile, we benchmark RAG-augmented foundation models, covering both LLMs and LMMs, together with a variety of retrievers.
We highlight two key observations: competition problems share similar concepts across years, and past problems capture not only the necessary physics knowledge from basic to advanced levels, but also valuable strategies for applying laws and formulas to novel scenarios. Importantly, PhoPile incorporates multimodal content—including diagrams, graphs, and equations—mirroring the real-world practice of physics problem solving. We further design an LLM-as-judge framework for evaluation. Our approach uses instructions and reference solutions to automatically score generated outputs, incorporating both step-wise and solution-level assessments to capture the richness of physics reasoning. This framework enables scalable evaluation that is otherwise infeasible with traditional script-based methods.


Our contributions are threefold:
\begin{itemize}
    \item We propose PhoPile, the first multimodal benchmark for evaluating retrieval-augmented physics reasoning. 
    \item We introduce a new LLM-as-judge evaluation framework tailored for physics, capable of handling diverse solution formats.  
    \item We conduct a comprehensive benchmark of 8 foundation models with 4 text-only retrievers and 3 multimodal retrievers, providing the first systematic study of RAG for physics reasoning.
\end{itemize}

\section{The PhoPile dataset}
\subsection{Overview}
PhOPile is structured to evaluate the performance of the RAG pipeline in the domain of high-level physics problem-solving. Therefore, we collect Physics Olympiad questions from various regions around the globe, including International Physics Olympiad (IPho), Asian Physics Olympiad (APhO), European Physics Olympiad (EuPhO), Nordic-Baltic Physics Olympiad (NBPhO), Romanian Master of Physics (RMPh), United States Physics Olympiad (USAPhO), and British Physics Olympiad (BPhO), all of which are publicly available.

Our data collection is motivated by the real-world practice of examinees reviewing past exam problems when preparing for future tests.  
Physics competition problems across years often share overlapping knowledge points, cover essential concepts from basic to advanced levels, and showcase diverse strategies for applying physical formulas.  
Organizing the dataset in this way provides rich references for tackling new problems by leveraging the knowledge and methods embedded in past competitions.

To evaluate foundation models with RAG in the context of physics, we organize the collected data as follows:
\begin{itemize}
    \item Evaluation Set: $390$ Olympiad-difficulty-level physics problems from 2019-2021, used to evaluate the model's performance on contemporary problems;
    \item Retrieval Corpus: $2,662$ problems before 2019, used by the retriever to provide context and reference for solving new problems.
\end{itemize}
The evaluation set is further divided into two subsets:
\textit{PhoPile-Test}, which contains 267 questions from 125 main problems in text-only form, and
\textit{PhoPile(V)-Test}, which consists of 123 questions from 77 main problems that include images either in the question statement or in the reference solutions.



\subsection{Data Collection}
In this section, we describe the process of collecting and standardizing physics competition problems into a unified format. A typical physics problem consists of textual descriptions, mathematical formulas, and images, and often exhibits hierarchical sub-question structures with multiple reference solutions. All finalized samples are stored in JSON format. 

\begin{figure*}[ht]
\centering
\begin{minipage}[b]{0.48\linewidth}
\centering
\begin{tabular}{lc}
\hline
\textbf{Evaluation Set} \\ \hline
w/ image & 117 \\ \hline
w/o image & 273 \\ \hline
\textbf{Retrieval Corpus} \\ \hline
w/ image & 879 \\ \hline
w/o image & 1,783 \\ \hline
\end{tabular}
\captionof{table}{Number of questions with or without images.}
\label{table:question_quantities}
\end{minipage}
\hfill
\begin{minipage}[b]{0.48\linewidth}
\centering
\includegraphics[width=0.9\linewidth]{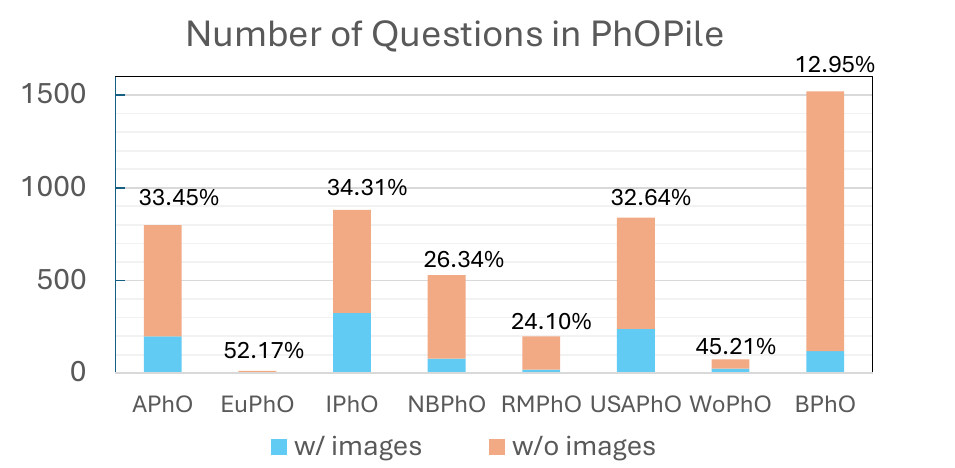}
\caption{Proportion of image-containing questions across different dataset sources.}
\label{fig:Oveview_img}
\end{minipage}
\end{figure*}

To construct PhoPile in a consistent and high-quality manner, we applied the following preprocessing steps:  \\
\textbf{(1) Text Cleaning.} We delete extraneous elements from the questions, like the historical background introductions, scoring criteria, and regulations or policies related to competitions. We filtered out certain LaTeX commands solely involved in adjusting the format, as they contribute nothing to the essence of the question. \\
\textbf{(2) Formula Representation in LaTeX.}  
As physics problems often involve mathematical formulas, we use LaTeX to represent the solutions with detailed information in plain text, inspired by the LaTeX-based representation adopted in MathPile~\cite{wang2024mathpilebilliontokenscalepretrainingcorpus}. Additionally, since the textual formulation of many physics problems is not directly accessible, we utilize the MathPix OCR tool\footnote{\url{https://mathpix.com}} to convert the content of images into LaTeX code, avoiding extensive manual effort while ensuring consistency. \\
\textbf{(3) Image Processing.} 32\% of problems in PhoPile involve images, a detailed breakdown of image usage is provided in Figure~\ref{table:question_quantities} and Figure~\ref{fig:Oveview_img}; we store them in the local repository and list the local URLs for the images associated with the questions and the solutions in PhoPile, respectively.
To highlight the position of the image appearing in the text, we mark $n$-th image as \verb|###img_n###|. 
Regarding the captions of these images, similar to MathVista~\cite{lu2024mathvista}, we omit the useless image labels (like `Figure~05' or `figure 1') and add the meaningful captions that contain crucial content relevant to the problem into the question description, such as `Figure 1: Isosceles glass prism with an apex angle of $90^\circ$'. \\
\textbf{(4) Hierarchical Question Structure.}  
Unlike mathematical datasets, physics problems often contain several sub-questions that may need to be answered in a specific order. We organize these sub-questions using Arabic numerals as indices. For the rest of the paper, we do not distinguish which main question the sub-questions belong to, except during evaluation. \\
\textbf{(5) Multiple Solutions.} 
For certain questions, the source files provide multiple solutions. They often appear as `Solution 2', `Another way to solve this problem'. For the completeness of the question and subsequent development, we also store them indexed by `solution 1', `solution 2', $etc$. 
Illustrative examples are provided in Appendix.


\subsection{Data Analysis} 

\begin{table*}[ht]
\small
    \centering
    \begin{tabular}{c|c c c c c c c|}
    \toprule
        Source & &\# Questions & \# Tokens & \# Tokens per q (Max/Min/Ave) & Years \\ \midrule
        \multirow{2}{*}{APhO} & evaluation set & 87 & 19,231 & 1,677/10/221 & 2019-2021\\
         & retrieval corpus & 502 & 85,949 & 1,708/7/171 & 2000-2018 \\ \midrule
        \multirow{2}{*}{EuPhO} & evaluation set & 15 & 2,489 & 3,951/20/902 & 2019-2021\\
         & retrieval corpus & 8 & 1,434 & 278/24/179 & 2017-2018 \\ \midrule
        \multirow{2}{*}{IPhO} & evaluation set & 93 & 16,871 & 854/15/181 & 2019-2021\\
         & retrieval corpus & 854 & 120,995 & 846/8/142 & 1967-2018 \\ \midrule
        \multirow{2}{*}{NBPhO} & evaluation set & 55 & 6,856 & 468/15/125 & 2019-2021\\
         & retrieval corpus & 374 & 36,400 & 569/5/97 & 2003-2018 \\ \midrule
        \multirow{2}{*}{RMPh} & evaluation set & 63 & 11,828 & 1,399/15/188 & 2019-2021\\
         & retrieval corpus & 132 & 15,251 & 792/9/116 & 2012-2018 \\ \midrule
        \multirow{2}{*}{USAPhO} & evaluation set & 77 & 6,373 & 262/10/82 & 2019-2021\\
         & retrieval corpus & 646 & 58,908 & 739/4/91 & 2007-2019 \\ \midrule
        \multirow{2}{*}{WoPhO} & evaluation set & 0 & - & - & -\\
         & retrieval corpus & 146 & 17,075 & 813/9/117 & 2011-2013 \\ \midrule
        \multirow{2}{*}{Total} & evaluation set & 390 & 63,648 & 3,951/10/ & 2019-2021 \\ 
         & retrieval corpus & 2,662 & 336,012 & 1,708/4/ & 1967 - 2019\\ \midrule
    \end{tabular}
    \caption{The token statistics of {\large P}hO{\large P}ile.}
    \label{token}
\end{table*}

\textbf{Statistics.} Token statistics are summarized in Table~\ref{token}.  
With the exception of EuPho, the majority of questions and solutions contain fewer than 500 tokens.  
This size is well within the context window of current popular LLMs such as Llama-2~\cite{touvron2023llama} and its variants, allowing for complete inference or training without the need for prompt pruning. 

\textbf{Images.} 
The overall image statistics are summarized in Table~\ref{table:question_quantities}. Images are widely present in PhoPile, appearing in both questions and solutions, with 33\% of problems containing at least one image.

Their distribution across different sources is shown in Figure~\ref{fig:Oveview_img}. These images play an essential role in conveying information such as experimental setups, physical systems, and data visualizations that cannot be fully captured by text alone.

In our PhoPile, about two-thirds of the solutions include images, ranging from curve plots that illustrate variable relationships to structural diagrams for force analysis, among other types.
Although most current LMMs are unable to generate images that precisely meet the requirements posed by the problems, we deliberately preserve these instances to encourage and support future research on multimodal reasoning in physics.

\section{Experiments}
\label{sec:experiments}
In this section, we introduce the RAG pipeline, describe the evaluation workflow, and present the experimental results of foundation models.

\subsection{RAG Pipeline}
The RAG pipeline comprises two main components—the retriever and the generator—and we further incorporate a reflection mechanism to enhance performance.

\textbf{Retriever.}  
Given an input query $q$, the retriever searches an external retrieval corpus $\mathcal{D}$ to find the most relevant problems and solutions. A scoring function $f(q, d_i)$ assigns relevance scores to each item $d_i \in \mathcal{D}$, and the top-$k$ items with the highest scores are selected:  
\[
\mathcal{R}(q) = \{d_1, d_2, \ldots, d_k\} \quad \text{where} \quad d_i \in \mathcal{D}.
\]
In PhoPile, $\mathcal{D}$ consists of 2,662 physics competition problems with reference answers collected before 2019. The retriever returns the top-$k$ relevant question–answer pairs $\mathcal{R}(q) = \{(q_i, a_i)\}_{i=1}^k$, where $q_i$ and $a_i$ denote the retrieved question and answer, respectively. The scoring function can thus be rewritten as $f(q, q_i)$.  

\begin{figure}
    \begin{tikzpicture}
    \node [draw, rounded corners,
         text width=\linewidth-24pt,    
         align=flush center, 
         inner sep=12 pt,
         fill=lightgray!40
    ]%
    {
    \begin{minipage}{00.9\linewidth}
    \footnotesize{
      Your task is to answer the physics questions. The mathematical formulas are provided in Latex code. There are some related questions and their answers you may find helpful. {\color{blue}{\textbackslash n}} Here are the examples:
    
    Question: \{{\color{orange}{Retrieved Question 1}}\} \\Reference answer: \{{\color{orange}{Reference Answer to Question 1}}\}
    
    Question: \{{\color{orange}{Retrieved Question 2}}\} \\ Reference answer: \{{\color{orange}{Reference Answer to Question 2}}\}{\color{blue}{\textbackslash n}} \\
    The question that you need to solve is: 
    {\color{blue}{\textbackslash n}}
    \{{\color{orange}{Question to be answered}}\}  {\color{blue}{\textbackslash n}}{\color{blue}{\textbackslash n}} \\
    Respond with the FINAL answer to the question to get a higher score as possible as you can, rather than only give directions or suggestions for solving the problem. Do NOT use the conditions in the example questions to solve the question.
    }
    \end{minipage}
     };
\end{tikzpicture}
\caption{
Instruction prompt template for the generator to answer the question.
}
\label{fig:prompt_generator}
\end{figure}
As there is no domain-specific retriever for physics, we adopt general-purpose retrieval methods. 
For text-only retrieval, we adopt several representative methods: (i) a sparse retriever, BM25~\cite{robertson2009probabilistic}; and (ii) dense retrievers, including Emb-cos (embedding model \textit{all-MiniLM-L6-v2}~\cite{wang2020minilm} with cosine similarity), Dragon+~\cite{lin2023train}, and Contriever~\cite{izacard2021unsupervised}.
For multimodal retrieval, we employ CLIP~\cite{radford2021learning}, ALIGN~\cite{jia2021scaling}, and VisualBERT~\cite{li2019visualbert} to obtain joint text–image embeddings, and use cosine similarity to identify the closest problems, which are likewise used as exemplars.

\textbf{Generator.}  
The generator takes the retrieved pairs $\mathcal{R}(q)$ together with the original query $q$ and produces a coherent, contextually grounded response $r$:  
\[
r = G(p(q, \mathcal{R}(q))),
\]
where $p$ denotes the prompt template that integrates the retrieved information, and $G$ is the generative model.  
An illustration of the prompt design is provided in Figure~\ref{fig:prompt_generator}.
 In our evaluation, we benchmark both text-based LLMs and vision–language LMMs, including Llama-2-13B, GPT-3.5, Llama-3-70B~\cite{grattafiori2024llama3herdmodels}, DeepSeek-Math~\cite{shao2024deepseekmath}, GPT-4, Gemini-Pro~\cite{Reid2024Gemini1U}, GPT-4V~\cite{OpenAI_gpt4v2023}, and Gemini-Pro-Vision.

\textbf{Reflection.} We use a reflection mechanism based on GPT-4. While retrieved questions and solutions can improve the generator’s performance, they may also introduce noise that misleads reasoning. One cause is the long input context after retrieval, which can dilute focus and increase distraction. The reflection mechanism can mitigate this issue. As illustrated in Figure~\ref{fig:reflection}, given a physics problem and two candidate answers—one generated with RAG and one without—the model is prompted to compare their relative quality and select the response it considers more accurate. This self-reflection step reduces the negative impact of noisy retrievals and enhances the robustness of the final output.

\textbf{Workflow.} Given a problem, we apply our RAG pipeline to solve it.  
For problems with a single question, we directly prompt the LLM to solve it while inserting the retrieved question and answers into the prompt. 
For problems with multiple sub-questions, we retrieve question and answer pairs according to the sub-question to be solved and incorporate the retrieved context into the query.  
The LLM is then prompted to generate a solution, which is appended to the query as additional context.  
This updated query is fed back to the LLM to solve the next sub-question.  
The process repeats until all sub-questions under the same main problem are answered, enabling the model to leverage both retrieved knowledge and its own prior reasoning history.

\begin{figure}
    \begin{tikzpicture}
    \node [draw, rounded corners,
         text width=\linewidth-24pt,    
         align=flush center, 
         inner sep=12 pt,
         fill=lightgray!40,
    ]%
    {
    \begin{minipage}{0.9\linewidth}
    \footnotesize{
      Your task is to choose the answer with a higher score of the given physics problem.{\color{blue}{\textbackslash n}}{\color{blue}{\textbackslash n}}
      
      Question: \{{\color{orange}{Question to be answered}}\} {\color{blue}{\textbackslash n}}
      
      Answer 1:  \{{\color{orange}{Candidate answer without RAG}}\}  {\color{blue}{\textbackslash n}}
      Answer 2: \{{\color{orange}{Candidate answer with RAG}}\}{\color{blue}{\textbackslash n}}{\color{blue}{\textbackslash n}}
    
    Please give a reason and output the final answer number in side ``\#\#'', for example, \#\#1\#\#.
    }
    \end{minipage}
     };
\end{tikzpicture}
\caption{
Instruction prompt template for reflection to choose the answer with or without RAG.
}
\label{fig:reflection}
\end{figure}


\subsection{Evaluation}



\begin{table}[ht]
\centering
\begin{tabular}{c|cccc}
\hline
$k$ & 0 & 1 & 2 & 3 \\
\hline
Accuracy (\%) & 37 & 49 & 73 & 87 \\
\hline
\end{tabular}
\caption{Accuracy of GPT-4 as a grader across different tolerance thresholds $k$.}
\label{GPT4_mark_acc}
\end{table}

\paragraph{Marking Scheme.}  
\label{sec:scoring}
Physics Olympiad problems typically require long chains of logical reasoning; therefore, scoring must account not only for the correctness of the final answer but also for the quality of intermediate reasoning steps.  
Instead of adopting the original competition-specific marking schemes, which assign varying maximum scores across problems, we standardize the scoring by setting the maximum score of each problem to ten points.  

To this end, we design an \emph{LLM-as-judge} evaluation framework tailored for physics, supporting both holistic and step-wise scoring.  
As illustrated in Figure~\ref{fig:evaluation_prompt}, GPT-4 is employed as the grader: it is prompted to compare candidate model solutions against the reference solution and assign a score between $0$ and $10$.  
 
To examine the reliability of GPT-4 as a grader, we conducted a human evaluation on the \textit{PhoPile-Test} set. Three experienced instructors who regularly provide training for the IPhO independently graded the model-generated solutions based on the same rubric that was provided to the LLMs for scoring. The final human score for each solution was obtained by averaging the three annotators’ scores. GPT-4 was prompted with the same rubric and asked to assign scores to the same set of solutions. We then compared GPT-4’s scores with the averaged human annotations under varying tolerance thresholds $k$, where $k$ denotes the maximum allowable difference between the GPT-4 score and the human-assigned score, as shown in Table~\ref{GPT4_mark_acc}. Results show that although exact agreement is not achieved, GPT-4 provides sufficiently consistent judgments to capture relative performance differences across models.

\begin{figure}
    \begin{tikzpicture}
    \node [draw, rounded corners,
         text width=\linewidth-24pt,    
         align=flush center, 
         inner sep=12 pt,
         fill=lightgray!40,
    ]%
    {
    \begin{minipage}{0.85\linewidth}
    \footnotesize{
    You are a professional physicist and you will grade answers provided by physics students by reference to standard answers. The full score is 10 points, and the minimum score is 0 points. If the student gives the final answer, full marks will be awarded directly. If the student does not give the final answer or the final answer is incorrect, please score based on the proportion of correct calculation steps given by the student. You only need to output a score number.
    
    Standard answer: {\color{orange} {\{Reference Answer\}}}
    
    Student answer: {\color{orange} \{Candidate Answer from Generator\}}
    }
    \end{minipage}
     };
\end{tikzpicture}
\caption{Prompt template for the evaluator to score the candidate answer, given the reference answer.
}
\label{fig:evaluation_prompt}
\end{figure}

\paragraph{Evaluation Metric.}
We report both the average score (AS) the LLMs and LMMs earn and their pass rate (PR) over \textit{PhoPile-Test} and \textit{PhoPile(V)-Test}. 
The average score is defined as,
\begin{equation}
\small
\textbf{AS} = \frac{\text{Total points gained by candidate model}}{\text{Number of questions} \times 10} \times 100\%.
\end{equation}

We regard the generator (LLMs and LMMs) as successfully passing a problem if they answer the problem correctly and earn a score of 10. Therefore pass rate is defined as,
\begin{equation}
\small
\mathbf{PR} = 100\% \cdot \frac{N_{\text{full-score}}}{N_{\text{total}}}.
\end{equation}

Intuitively, the pass rate and average score reflect the performance of the retriever and generator in a coupled manner. The higher the value of the pass rate and average score, the better the performance of the retriever and generator.

\paragraph{Baselines.}


For the \textit{PhoPile-Test} subset containing text-only questions, we evaluate strong publicly available instruction-tuned LLMs, including LLaMA-3, as well as proprietary models trained with private data, such as GPT-3.5, GPT-4, and Gemini-Pro. We additionally evaluate DeepSeekMath-7B, which is trained with math-related tokens and achieves comparable performance of GPT-4.

For the \textit{PhoPile(V)-Test} subset with images, we use LMMs including GPT-4V and Gemini-Pro-Vision. By default, we evaluate the zero-shot CoT~\cite{wei2022chain} performance for the baselines.
To evaluate RAG's effectiveness in solving Olympic-level mathematical physics problems, we experiment with the different RAG methods, where a model generates output given the input question prepended with the top-K retrieved examples.

In addition, we finetune several open-source models on the retrieval corpus, which consists of question–answer pairs.  
The models include Mistral-7B-v0.3, Phi-3.5-mini, LLaMA-3-8B, and Mathstral-7B-v0.1.  
We denote the fine-tuned versions with the suffix `-FT', for example: Mistral-7B-v0.3-FT, Phi-3.5-mini-FT, LLaMA-3-8B-FT, and Mathstral-7B-v0.1-FT.


\subsection{Main Results}

\begin{table*}[ht]
    \centering
    \small
    \begin{tabular}{
        >{\centering\arraybackslash}p{3.4cm}|
        >{\centering\arraybackslash}p{0.7cm}|
        >{\centering\arraybackslash}p{1.4cm}
        >{\centering\arraybackslash}p{1.4cm}
        >{\centering\arraybackslash}p{1.4cm}
        >{\centering\arraybackslash}p{1.4cm}
        >{\centering\arraybackslash}p{1.4cm}}
    \toprule
    Model & Input & w/o RAG & Emb-cos & BM25 & Dragon+ & Contriever \\ 
     &  & PR(AS)& PR(AS) & PR(AS) & PR(AS) & PR(AS) \\ \midrule
    \multicolumn{4}{l}{On \textit{PhoPile-Test}}  \\ \midrule 
    Llama-3-70B & \textit{T} & 10.51(1.34) &5.4(1.84) &19.07(4.86) &13.62(4.83) &10.28(4.65) \\
    Llama-3-70B w/ Reflection & \textit{T} & 10.51(1.34) & 19.38(4.35)& 19.38(4.35)  & 14.51(4.81)  &  10.80(4.60) \\ 
    \hline
    DeepSeek-Math  &\textit{T} & 4.10(0.64) & 2.06 (0.27) &2.06(0.29) &2.06(5.93) &3.08(0.38) \\
    DeepSeek-Math w/ Reflection &\textit{T}  & 4.10(0.64)   & 16.95(2.85) &3.59 (0.55) & 2.83(6.085) &3.37 (0.54) \\
    \hline
    GPT-3.5     & \textit{T} & 7.95(4.12) &8.72(4.02) &8.23(3.84) &10(3.75) &7.69(3.91)\\ 
    GPT-3.5  w/ Reflection   & \textit{T} & 7.95(4.12) & 11.79(4.9)  & 9.23(4.37) & 10.26(4.25) &  8.46(43.21) \\ 
    \hline
    Gemini-Pro     & \textit{T} & 17.18(5.30) &16.15(4.91) & 15.90(4.93) &16.41(5.69) & \textbf{30.51(5.19)} \\
    Gemini-Pro w/ Reflection    & \textit{T} & 17.18(5.30)  & 21.54(5.72) & 20.51(5.56) & 18.72(5.65) & 19.74(5.49)\\
    \hline
    GPT-4      & \textit{T} & 26.41(6.27) &24.10(5.71) &25.19(5.92) &25.71(5.91) &25.19(5.82) \\ 
    GPT-4 w/ Reflection    & \textit{T}  & 26.41(6.27) & 27.92(6.22)  & 27.69(6.37) & 28.46(6.34) & 26.99(6.23)  \\
    \midrule
    
    Mistral-7B-v0.3-FT & \textit{T} & 1.47(2.10) & 22.64(5.38) & 20.90(4.10)& 25.28(4.62) & 23.28(6.15) \\ 
    Phi-3.5-mini-FT & \textit{T} & 2.56(1.95) & 18.00(7.18) & 16.25(6.43) & 20.31(7.95) & 21.44(7.46) \\ 
    Llama-3-8B-FT  & \textit{T} & 5.86(2.17) & 28.31(5.90) & 26.44(5.38) & 27.46(5.91) & 25.39(6.19) \\ 
    Mathstral-7B-v0.1-FT & \textit{T} & 6.62(2.84) & 27.17(5.91) & \textbf{29.02(9.28)} & 28.90(8.21) & 27.66(8.74) \\ \hline
    
    \multicolumn{4}{l}{On \textit{PhoPile(V)-Test}}   \\ \midrule
    Gemini-Pro-V & \textit{T, I} & 12.82(5.09) & 17.95(5.24) & 12.82(4.78)& 12.88(4.858) & 14.96(5.04)\\
    Gemini-Pro-V w/ Reflection  & \textit{T, I} & 12.82(5.09)  & 19.23(5.24) & 16.67(5.05)& 15.38(4.83) & 17.09(5.06) \\
    \hline
    GPT-4V & \textit{T, I} & 21.79(6.26)& 20.51(5.43)&7.69(2.50) & 21.46(5.53)& 21.79(5.65) \\ 
    GPT-4V w/ Reflection & \textit{T, I} & 21.79(6.26) & \textbf{21.89(6.09)} & 19.31(5.25) & 21.03(6.16) &21.46(6.20)  \\
    \bottomrule
\end{tabular}
    \caption{Evaluation results on PhoPile.  
    \textit{Input:} \textit{T} = question text only; \textit{T, I} = question text with images.  
    Values indicate pass rates (PR) in percentages, with average scores (AS) in parentheses.  
    Bold values denote the best performance.  
    All retrievers are text-only.}
    \label{tab:main_results}
\end{table*}
\begin{table*}[h]
    \centering
    \small
    \begin{tabular}{
        >{\centering\arraybackslash}p{4cm}|
        >{\centering\arraybackslash}p{0.7cm}|
        >{\centering\arraybackslash}p{1.4cm}
        >{\centering\arraybackslash}p{1.4cm}
        >{\centering\arraybackslash}p{1.4cm}
        >{\centering\arraybackslash}p{1.4cm}}
    \toprule
    Model & Input & w/o RAG & CLIP & VisualBERT & ALIGN \\ 
     &  & PR(AS) & PR(AS) & PR(AS) & PR(AS) \\ \midrule

    Gemini-Pro-V & \textit{T, I} & 12.82(5.09) & 17.48(4.99) & 13.59(3.42) & 14.56(5.88) \\

    Gemini-Pro-V w/ Reflection & \textit{T, I} & 12.82(5.09) & 14.56(5.12) & \textbf{17.48(5.28)} & 15.53(5.35) \\ \hline

    GPT-4V & \textit{T, I} & 21.79(6.26) & \textbf{30.10(6.20)} & 24.27(5.80) & 15.53(5.79) \\ 

    GPT-4V w/ Reflection & \textit{T, I} & 21.79(6.26) & 26.41(5.99) & 22.33(5.58) & 23.30(5.71) \\ 
    \bottomrule
\end{tabular}
\caption{Evaluation results on \textit{PhoPile(V)-Test} with multimodal retrieval. 
Bold values denote the best-performing retriever for each model.}
\label{tab:main_results_multi_modality}
\end{table*}

We report the evaluation results of the foundation models with text-only retrieval under the 2-shot setting in Table~\ref{tab:main_results}, including LLMs on \textit{PhoPile-Test} and LMMs on \textit{PhoPile(V)-Test}. RAG provides some insights to aid the generators' physical reasoning. For example, Gemini-Pro combined with Contriever improves substantially, from 17.18\% to 30.51\%, while LLaMA-3-70B with BM25 increases from 10.51\% to 19.07\%. However, not all retrievers yield positive effects.  
In many cases, performance decreases with RAG, primarily due to noise and irrelevant content introduced by the retrieved examples.  For LMMs with RAG, GPT-4 consistently outperforms Gemini in both pass rate and average score.  
With the incorporation of RAG, Gemini-Pro improves from 12.82\% to 17.95\%.  
We also investigate reflection in the RAG framework as a means to mitigate the negative impact of retrieved questions. This mechanism yields noticeable performance improvements.

Interestingly, among open-source models, we observe substantial improvements after fine-tuning even for those with fewer than 8 billion parameters. Their overall performance increased by factors ranging from 5 to 17. The strongest model, Mathstral, achieved an accuracy of 29.02, which is already comparable to the best closed-source result of 30.51.

\begin{table*}[ht]
    \centering
    \footnotesize
    \begin{tabular}{
        >{\centering\arraybackslash}p{1.8cm}|
        >{\centering\arraybackslash}p{1.0cm}|
        >{\centering\arraybackslash}p{1.5cm}|
        >{\centering\arraybackslash}p{1.5cm}|
        >{\centering\arraybackslash}p{1.5cm}|
        >{\centering\arraybackslash}p{1.5cm}}
    \toprule
    Model & \#Shots & Emb-cos & BM25 & Dragon+ & Contriever \\ 
     &  & PR(AS) & PR(AS) & PR(AS) & PR(AS) \\ \midrule
    \multirow{3}{*}{GPT-3.5}   
        & 1 & 8.97(3.88) & 6.92(3.65) & 9.74(3.87) & 0.77(0.62) \\ 
        & 2 & 8.72(4.02) & 8.23(3.84) & 10.00(3.75) & 7.69(3.91) \\ 
        & 3 & 9.74(3.90) & 6.41(3.77) & 7.44(3.70) & 7.71(3.88) \\ \midrule
    \multirow{3}{*}{GPT-4}  
        & 1 & 26.74(6.00) & 22.82(5.70) & 26.41(6.01) & 28.97(6.10)  \\ 
        & 2 & 24.10(5.71) & 25.19(5.92) & 25.71(5.91) & 25.19(5.82) \\
        & 3 & 25.90(6.01) & 22.56(5.65) & 22.37(5.89) & 24.62(5.91) \\ 
    \bottomrule
    \end{tabular}
    \caption{Evaluation results of GPT-3.5 and GPT-4 with different numbers of retrieved examples. 
    The rows correspond to different numbers of shots (1, 2, and 3), with values in parentheses indicating standard deviations. 
    }

    \label{tab:rag_shot}
\end{table*}

The results of multimodal retrieval based on joint text–image embeddings are presented in Table~\ref{tab:main_results_multi_modality}. Both Gemini-Pro-V and GPT-4V show improvements after applying multimodal RAG, though the extent varies across retrievers. GPT-4V benefits most from CLIP, reaching a pass rate of 30.10\%, while Gemini-Pro-V gains more from VisualBERT, where reflection boosts its performance to 17.48\%. These results demonstrate that the choice of multimodal retriever significantly impacts performance, with CLIP particularly effective for GPT-4V and VisualBERT more favorable for Gemini-Pro-V.

\paragraph{Error analysis.}
\label{sec:error}

In many cases, retrieved examples had a negative impact on performance, primarily due to the following reasons: 1) The general retriever was not effectively applied to physics problems, as retriever specific to physics may consider the questions that using the same theorem as the top-k relevant ones, instead of those with highest semantic similarity. Therefore, it highlights the significance of establishing domain-specific retrievers.
2) The format in retrieved questions misleads the candidate models' answering. The retrieved questions and their reference answer may provide guidance answers instead of directly answering the question. Therefore, the foundation models may refuse to answer the final answer directly and answer with some guidance for the question.
3) Additionally, some wrong answers arise from using conditions in the retrieved questions as if they were the known conditions in the current question, demonstrating the significance of noise robustness in foundation models with RAG. 
Please refer to Appendix for the examples.






\subsection{Ablation}  


Table~\ref{tab:rag_shot} reports the evaluation results of GPT-3.5 and GPT-4 using different retrieval methods: Emb-cos, BM25, Dragon+, and Contriever, with varying numbers of retrieved examples: 1-shot, 2-shot, and 3-shot.  
The results show that the top-1 retrieved example from Contriever can negatively affect the performance of GPT-3.5, whereas Emb-cos and Dragon+ provide more useful examples, particularly in the 1-shot and 2-shot settings.


\section{Related Work}

RAG~\cite{Chen_Lin_Han_Sun_2024, zhang2024retrievalqa} has recently drawn significant attention in complimenting the domain-specific expertise for LLMs ~\cite{zhangruag},  or constructing demonstrations for in-context learning (ICL)~\cite{poesia2022synchromesh, agrawal-etal-2023-context, liu-etal-2022-makes, hu-etal-2022-context, li-etal-2023-unified}, thus serves as a natural way to enhance foundation model's capability of physic reasoning by integrating external knowledge sources.  Let's consider a high school student preparing for an exam; it is natural for the student to review past exam questions to find similar types of problems to practice. These questions can provide similar problem-solving approaches and relevant knowledge applications, much like how RAG can retrieve and incorporate pertinent information and similar demonstrations to enhance the reasoning and accuracy of LLMs, resulting in more informed and contextually relevant responses.

In recent years, LLMs have developed very rapidly, providing great convenience for people's needs in all aspects of life~\cite{wang2025large, liu2023aliu2024agentbenchgentbench, hu2025agents,zhong2023rspt,schipper2025pillagerbench}. These models, like GPT-3, GPT-4 and Gemini have already shown great performance in terms of accuracy, interpretability, and multimodality, similarly as general LLMs, they show outstandingly high performance of natural science QA and mathematical reasoning. Meanwhile, a range of excellent open source models, including T5~\cite{2020t5}, GPT-2~\cite{radford2019language} and Llama-2, is available for researchers to enhance further, by training them on a specialized dataset to attain superior capabilities compared to generalized models~\cite{magister2023teaching, shridhar2022distilling, zhang2025distilldatarewardssmaller}. Consequently, a series of outstanding open source models that are specifically trained and fine-tuned on math have emerged, such as DeepSeekMath, Llema and Goat~\cite{Liu2023GoatFL}. Additionally, there are also a few models focusing on formal proof such as LeanDojo~\cite{yang2023leandojo}; these are models trained on math-specialized corpus or datasets. However, in the expansive domain of mathematics, the multitude of sub-disciplines presents a significant challenge for models with constrained parameters to adequately address comprehensive mathematical problems. Studies like Boosting LLM Reasoning~\cite{Huang2023BoostingLR} and LeanDojo use a retrieval-augmented approach to improve the accuracy of mathematical problem-solving. It is noteworthy that research at the intersection of linguistics and natural sciences remains relatively scarce. Scholars have placed a greater emphasis on mathematical reasoning.


Models which demonstrate excellent performance on mathematical ability are inseparable from high-quality datasets and corpus such as Mathpile, proof-pile-2, MiniF2F, MATH, GSM8K, MLFMF~\cite{bauer2023mlfmf},
MathOdyssey and the corpus proposed by DeepSeekMath. The aforementioned datasets consist solely of textual data; however, it is commonly understood that the interpretation of mathematical problems often requires the analysis of images. Consequently, MathVista introduced a specialized image-based mathematical dataset and conducted evaluations of models such as GPT-4, GPT-3.5, Claude-2~\cite{anthropic2023claude2}, and mPLUG-Owl-Llama~\cite{Ye2023mPLUGOwlME} from various perspectives: purely textual input, text with captions and image OCR (Augmented-LLMs), and multimodal analysis. However, there is a noticeable paucity of specialized research linking LLMs with the discipline of physics. The relevant work in this area is confined to a minimal subset of physics-related data within certain natural science datasets, such as SciQ Dataset, ScienceQA, C-eval~\cite{Huang2023CEvalAM}, E-EVAL~\cite{Hou2024EEVALAC}, and TheoremQA. 


\section{Conclusion}  
In this work, we present \textbf{PhoPile}, a benchmark designed to comprehensively evaluate the ability of foundation models to perform physics reasoning with retrieval-augmented generation (RAG) across both text-only and image-based questions.  
We benchmark a range of mainstream foundation models, including both large language models (LLMs) and large multimodal models (LMMs), together with multiple retrievers.  
To ensure robust evaluation, we introduce an LLM-as-judge framework capable of assessing diverse solution formats.  
Our results demonstrate that combining physics corpora with retrieval can improve performance, while also revealing challenges that motivate further research in retrieval-augmented physics reasoning.

\section*{Limitations}
\label{sec:limitations}
Our retrieval corpus is limited in scale due to constraints in data acquisition, including restricted access to diverse sources and practical challenges in collection and integration.  
As future work, we plan to incorporate multimodal cross-referencing, which would enable richer interactions between text, formulas, and images, and has the potential to further improve the accuracy and robustness of foundation models for physics reasoning.  

\bibliography{custom}

\clearpage
\onecolumn 
\appendix

\section*{Appendix}

\begin{enumerate}
  \item \textbf{Question Examples in {\large P}hO{\large P}ile} (Sec.~\ref{example_json})\\
  Dataset indexing scheme (``question\_number'', ``sub\_question\_number'', ``sub\_sub\_question\_number'') and textual normalization of in-question indices.\\
  \emph{Figure:} Examples from the dataset (Figure~\ref{fig:examples}).

  \item \textbf{Supplementary Experiments and Examples for GPT-4 Scoring} (Sec.~\ref{sec: Supplementary})\\
  Scoring rubric and qualitative marking cases for GPT-4 as evaluator.\\
  \emph{Table:} Candidate-answer categories and expected scores (Table~\ref{Validation_table}).\\
  \emph{Figures:} 10/5/0-score answer exemplars (Figures~\ref{Prompt_example_e1}--\ref{Prompt_example_e3}).

  \item \textbf{Prompt Examples} (Sec.~\ref{sec:prompts})\\
  Prompt formats for pure-text inference and 1-shot RAG.\\
  \emph{Figures:} Pure text prompt (Figure~\ref{Prompt_example}); 1-shot RAG prompt (Figure~\ref{rag_Prompt_example}).

  \item \textbf{Gemini Example} 
  (Sec.~\ref{sec:Gemini Example})\\
  One worked Gemini inference case with 1-shot RAG.\\
  \emph{Figure:} Gemini output example (Figure~\ref{Gemini_Example}). 

  \item \textbf{Multimodality Example} 
  (Sec.~\ref{sec:Multimodality})\\
  Vision-augmented physics problem; standard answer; GPT-4-V vs Gemini-Pro-Vision responses and scores.\\
  \emph{Figure:} The image from the problem (Figure~\ref{fig:image1}). 

  \item \textbf{Runtime Analysis} (Sec.~\ref{app:runtime_details})\\
  Retrieval vs retrieval+embedding timing across methods.\\
  \emph{Table:} RAG performance (Table~\ref{tab:rag_performance}).

  \item \textbf{Examples of Answer Errors} (Sec.~\ref{app:error_analysis})\\
  \begin{enumerate}
    \item \emph{Error Type 1:} Gives guidelines instead of a direct answer (projectile problem).
    \item \emph{Error Type 2:} Misguided by retrieved examples (thermoacoustic temperature amplitude).
    \item \emph{Error Type 3:} Wrong retrieval (dipole torque/power vs unrelated contexts).
  \end{enumerate}

  \item \textbf{Additional Results} (Sec.~\ref{app:finegain_results})\\
  Difficulty distribution; RAG-driven zero\,$\to$\,non-zero improvements; average scores by difficulty.\\
  \emph{Tables:} Difficulty distribution (Table~\ref{tab:difficulty_distribution}); average scores by difficulty (Table~\ref{scores}); questions improved from 0 to non-0 after RAG (Table~\ref{tab:model_improvement}).
\end{enumerate}

\clearpage
\section{Question Examples in {\large P}hO{\large P}ile}
\label{example_json}
To fully record the details of the questions, we create `question\_number', `sub\_question\_number', and `sub\_sub\_question\_number' which stand for question number, first order sub-question number and second order sub-question number in our dataset to facilitate distinction. Furthermore, within the text of the questions, we replace the original question index, which typically consist of Arabic numerals, English letters, Roman numerals, etc, with Arabic number in `\texttt{sub\_question\_number}' or `\texttt{sub\_sub\_question\_number}'. We provide examples in Figure~\ref{fig:examples}.

\begin{figure*}
    \begin{tikzpicture}
    \node [draw, rounded corners,
         text width=\linewidth-24pt,    
         align=flush center, 
         inner sep=12 pt,
         fill=lightgray!40,
    ]%
    {
    \begin{minipage}{00.9\linewidth}
    \footnotesize{
        {\color{blue} Example 1:} \\
        ``index'': 2882, \\
        ``problem'': "A metal ring of radius $R$ is made out of a wire of cross section aria $s$; the resistivity of the wire's material is given by $\rho$. The ring is place in a vertical plane. A small magnetic needle is placed in the center of the ring. The magnetic needle can rotate freely around the vertical axis which passes through the center of the ring. Due to mechanical inertia, the needle points in the direction of the average horizontal magnetic field at the center of the ring. When the ring is motionless in the reference frame of the Earth, the magnetic needle indicates the direction of the horizontal component of Earth's magnetic field. When the ring is rotating around its vertical axis with a constant angular velocity, $\omega$, the magnetic needle deviates, in horizontal plane, from this direction by an angle $\alpha$. 1.a. Find the expression of angle of deviation, $\alpha$ of the magnetic needle. Write your expression for $\alpha$ as a function of $s, \omega, \rho$ and of the magnetic permeability $\mu_0$.'', \\
        ``question\_number'': 978, \\
        ``sub\_question\_number'': 1, \\
        ``sub\_sub\_question\_number'': null, \\
        ``source'': "RMPh'', \\
        ``year'': 2021, \\
        ``solution'': ``For: $ \vec{B}_{\text {Earth }}=B_h \cdot \hat{e}_x+B_v \cdot \hat{e}_z $ $B_h$ - horizontal component of Earth's magnetic field $B_v$ - vertical component of Earth's magnetic field formula of surface vector $\vec{S}$ for the ring $ \vec{S}=\pi \cdot R^2 \cdot\left(\cos (\omega \cdot t) \cdot \hat{e}_x+\sin (\omega \cdot t) \cdot \hat{e}_y\right) $ expression of magnetic flux $\phi$ of Earth's magnetic field through ring's surface ...'', \\
        ``imgQ'': null, \\
        ``imgA'': null \\
        
        {\color{blue} Example 2:} \\
        ``index'': 2883, \\
        ``problem'': ``A thin ring of mass $m$, radius $r_0$ and inductance $L$ is maintained in a horizontal plane above a cylindrical magnetic bar which is placed vertically (see \#\#\#img\_1\#\#\#). The vertical axis of symmetry of the cylindrical magnetic bar is aligned with the center of the ring. The magnetic field due to the cylindrical magnet is shown in \#\#\#img\_1\#\#\# and each of its components are given by: the radial component is given by $B_r=B_0 \cdot \beta \cdot r$, while the vertical component is given by $B_z=B_0 \cdot(1-\alpha \cdot z) . B_0, \alpha, \beta$ are all positive constant with appropriate dimensions, while $z$ and $r$ denotes the vertical and, respectively, the radial coordinate of the system. Initially, there is no electric current passing through the ring and it is kept fixed above the magnet. It is then allowed to fall due to the gravitational pull of the Earth, given by the gravitational acceleration, $g$. During the fall the ring will still be in a horizontal plane and will have the same vertical axis.  Answer the following questions and write your results as a function of the variables specified above. Derive the equations of motion for the ring in the reference frame specified in \#\#\#img\_1\#\#\#.'', \\
        ``question\_number'': 978, \\
        ``sub\_question\_number'': 2, \\
        ``sub\_sub\_question\_number'': 1, \\
        ``source'': "RMPh'', \\
        ``year'': 2021, \\ 
        ``solution'':  ``For: expression of magnetic flux through ring's surface $\Phi=B_z \cdot \pi \cdot r_0^2+L \cdot I$ $ 0=R \cdot I=\frac{d \Phi}{d t} $ - voltage drop on superconducting ring is zero - magnetic flux inside the ring is constant $ \Phi=B_0 \cdot(1-\alpha \cdot z) \cdot \pi \cdot r_0^2+L \cdot I=\text { constant } $  Initial conditions $\left\{\begin{array}{l}z(t=0)=0 \\ l(t=0)=0\end{array} \quad\right.$ constant $=B_0 \cdot \pi \cdot r_0^2$ expression of the intensity of electric current through the ring $I=\frac{B_0}{L} \cdot \alpha \cdot \pi \cdot r_0^2 \cdot z$ radial component of the force of interaction is zero - because of symmetry vertical component of the force of interaction $F_z=-\frac{2 \pi^2 \cdot \alpha \cdot \beta \cdot B_0^2 \cdot r_0^4}{L} \cdot z$ elastic constant $k=\frac{2 \pi^2 \cdot \alpha \cdot \beta \cdot B_0^2 \cdot r_0^4}{L}$ equations of motion for the ring $m \cdot \frac{d^2 z}{d t^2}+k \cdot z=-m \cdot g$ general solution of the equations of motion for the ring ...'', \\
        ``imgQ'': [
            "18.png"
        ], \\
        ``imgA'': null \\
    }
    \end{minipage}
     };
\end{tikzpicture}
\caption{Examples from the dataset.}
\label{fig:examples}
\end{figure*}

\section{Supplementary Experiments and Examples for GPT-4 Scoring}
\label{sec: Supplementary}

As shown in Table~\ref{Validation_table}, we verify the GPT4's capability on marking the candidate answers by foundation models, given the reference solution.

\begin{figure}
    \centering
    \includegraphics[width=0.82\textwidth]{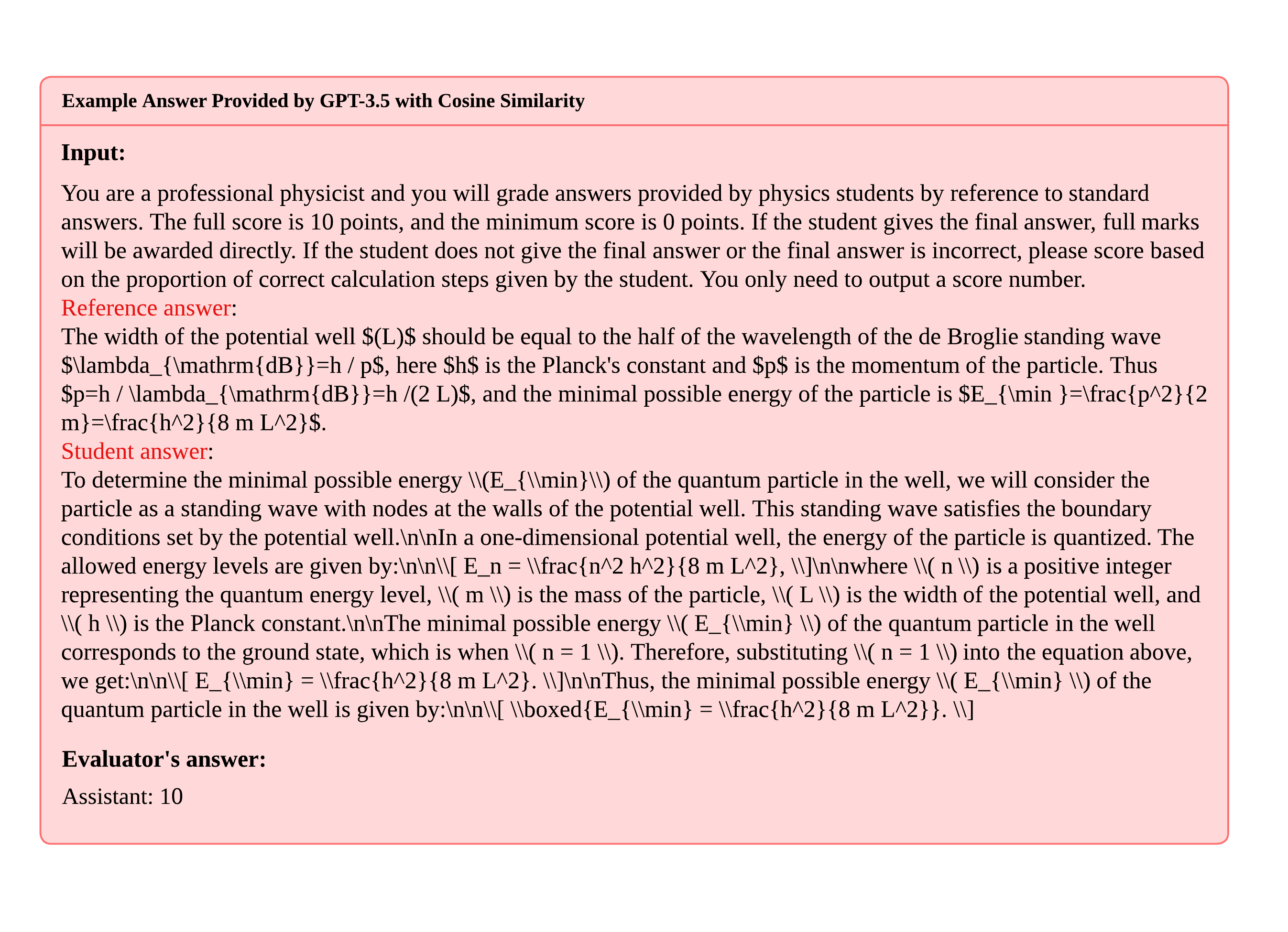} 
    \caption{A 10 score example answer provided by GPT-3.5 with cosine similarity.}
    \label{Prompt_example_e1}
\end{figure}

\begin{figure}
    \centering
    \includegraphics[width=0.82\textwidth]{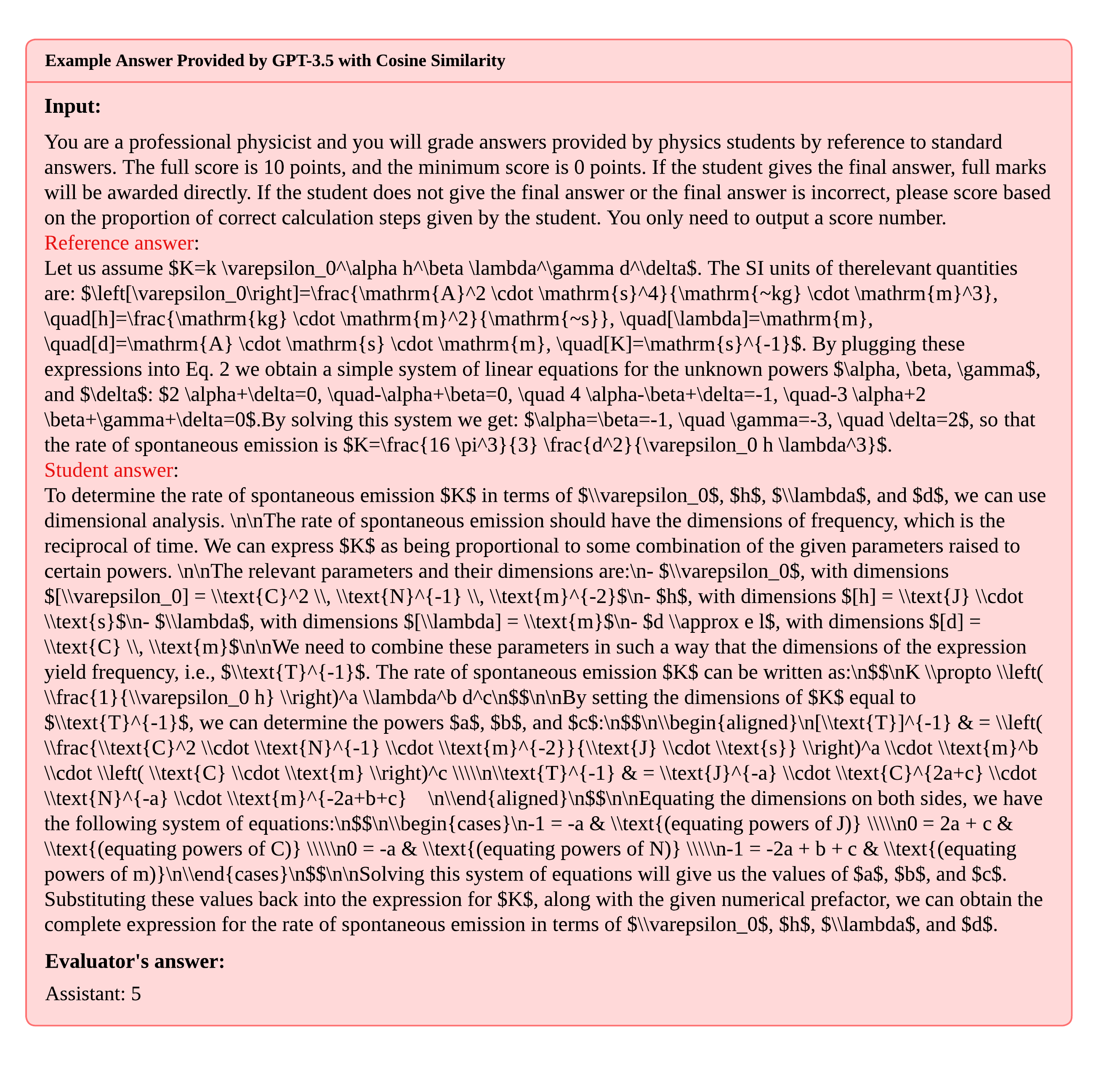} 
    \caption{A 5 score example answer provided by GPT-3.5 with cosine similarity.}
    \label{Prompt_example_e2}
\end{figure}

\begin{figure}
    \centering
    \includegraphics[width=0.82\textwidth]{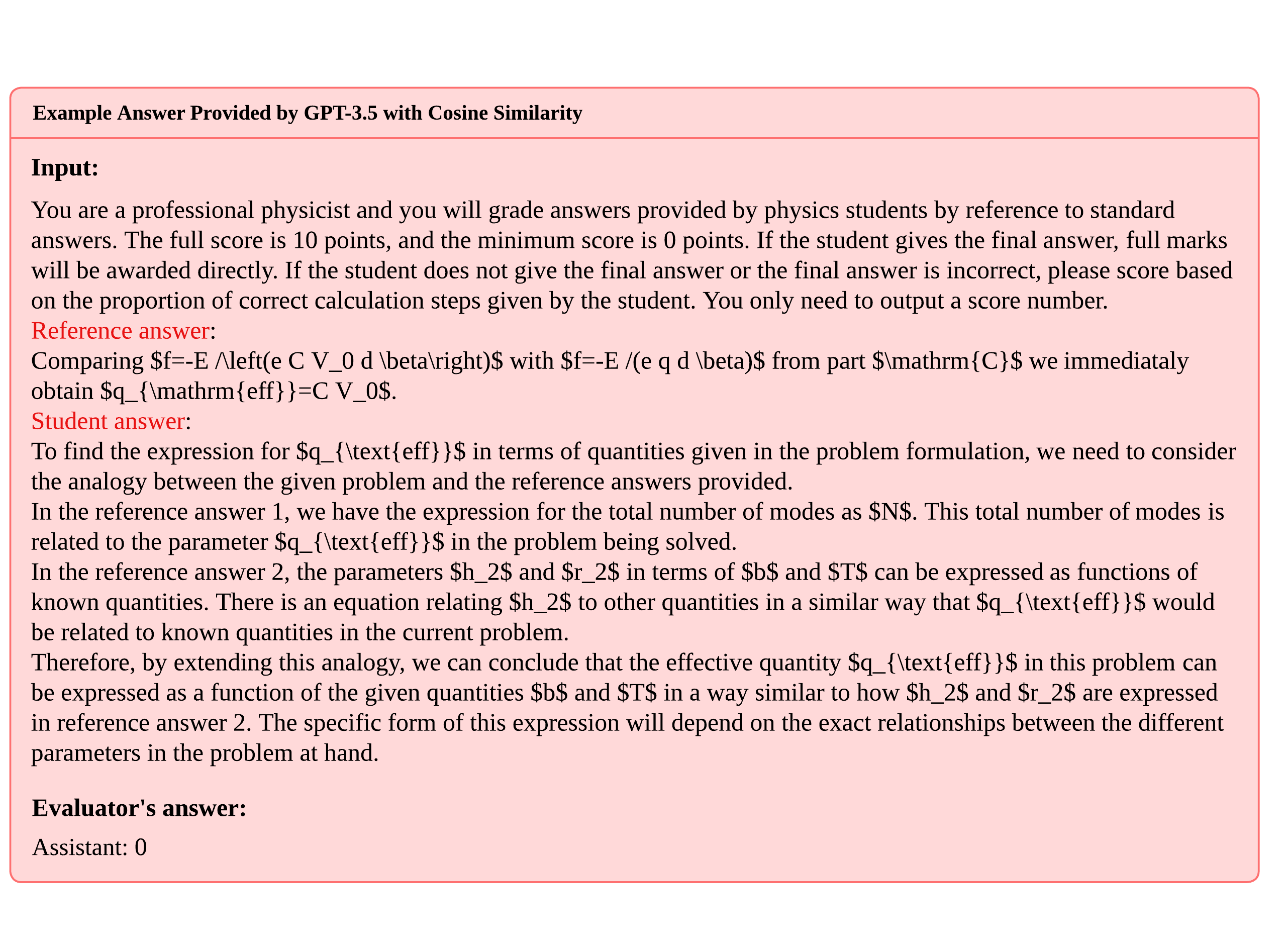} 
    \caption{A 0 score example answer provided by GPT-3.5 with cosine similarity.}
    \label{Prompt_example_e3}
\end{figure}

Figure \ref{Prompt_example_e1}, Figure \ref{Prompt_example_e2} and Figure \ref{Prompt_example_e3} are examples of evaluator marking three different types of answers.

\clearpage
\begin{table}
    \centering
    \begin{tabular}{p{0.8\linewidth}|>{\centering\arraybackslash}p{0.1\linewidth}}
    \toprule
        Candidate Answer  &  Output \\ \midrule
        Fully correct solution &  10 \\ 
        Fully incorrect solution but with the correct final answer  &  0 \\ 
        A completely wrong solution  &  0 \\ 
        Half of the fully correct solution &  1-9 \\ 
       Fully correct solution but with a wrong final answer  &  9 \\ 
        Correct answers in various type (normal format, Latex format, or without units) &  10 \\ 
        \toprule
    \end{tabular}
    \caption{By inputting the candidate solution to be marked and reference answer to GPT4, we can have detailed and based-on-step scores without human intervention. 
    }
    \label{Validation_table}
\end{table}

\clearpage
\clearpage

\clearpage

\clearpage
\section{Prompt Examples}
\label{sec:prompts}
\begin{figure}[hp]
    \centering
    \includegraphics[width=0.82\textwidth]{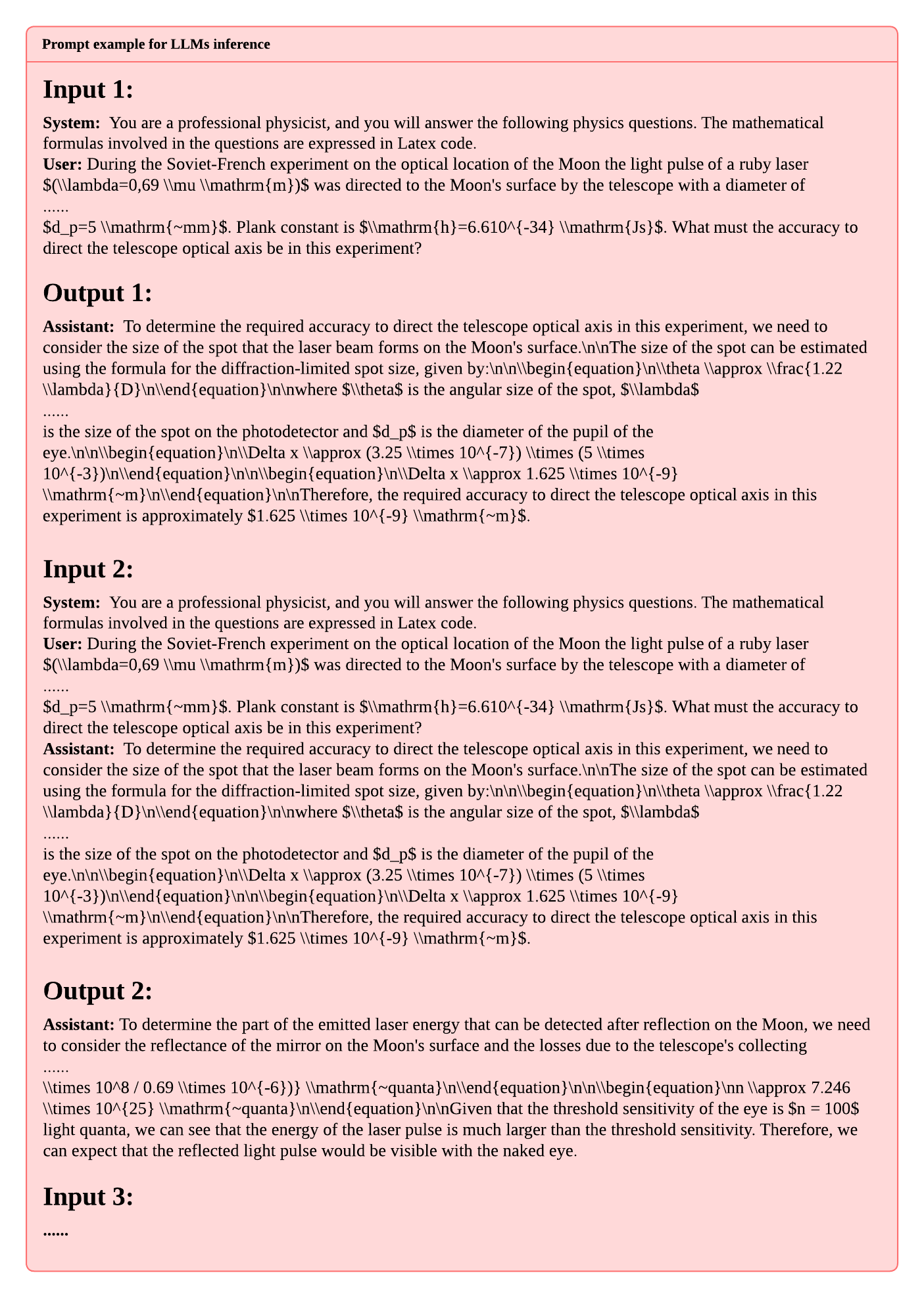} 
    \caption{Prompt example of pure text inference}
    \label{Prompt_example}
\end{figure}

\begin{figure}[hp]
    \centering
    \includegraphics[width=0.82\textwidth]{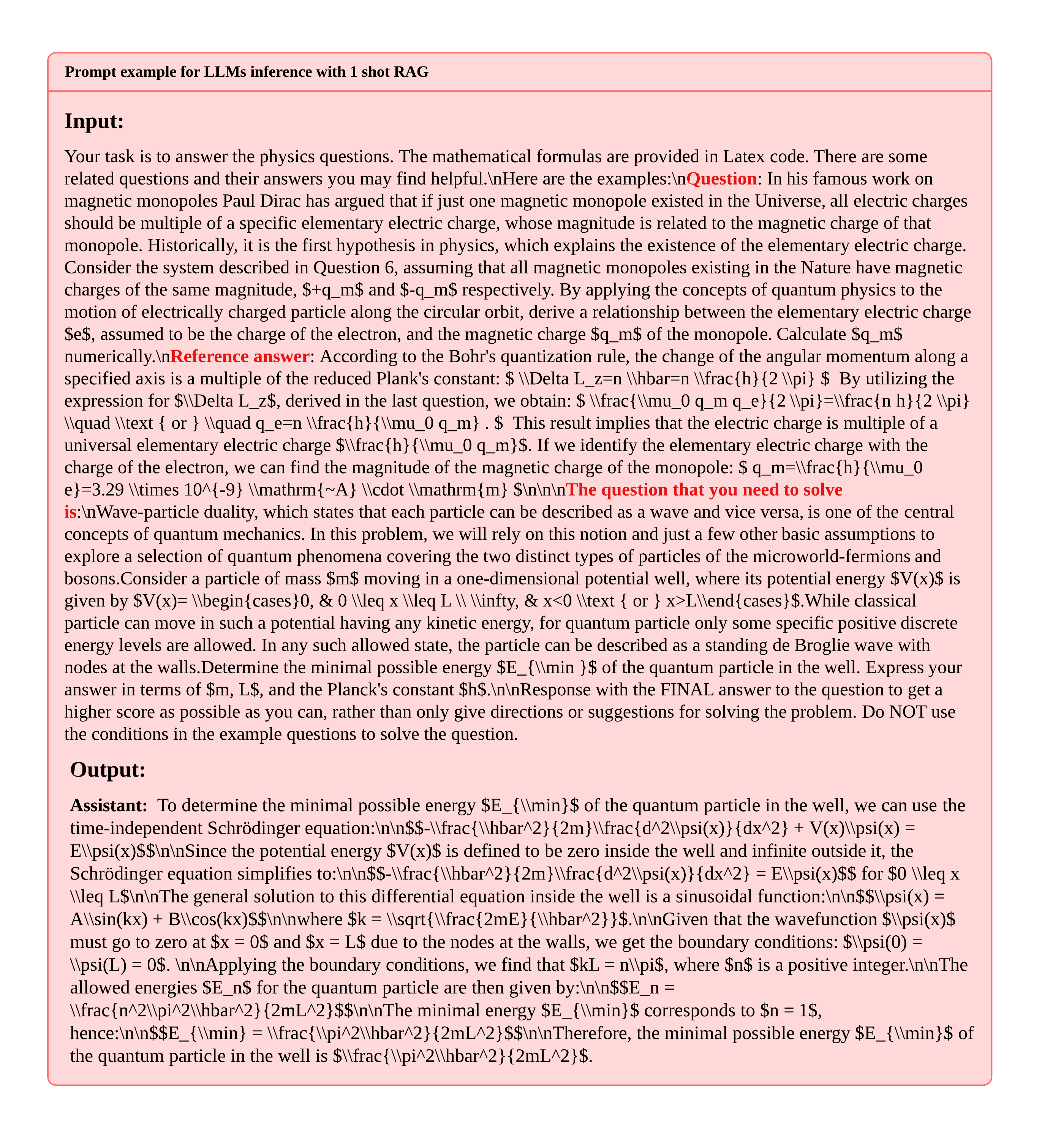} 
    \caption{Prompt example of pure text inference with 1-shot RAG}
    \label{rag_Prompt_example}
\end{figure}

In this section, we present the prompt examples used to guide LLMs in solving physical questions and evaluating solutions. 
Figure~\ref{Prompt_example} shows a prompt example of pure text inference, while Figure~\ref{rag_Prompt_example} illustrates a prompt with 1-shot RAG.

\clearpage
\section{Gemini Example}
\label{sec:Gemini Example}
\begin{figure}[hp]
    \centering
    \includegraphics[width=0.82\textwidth]{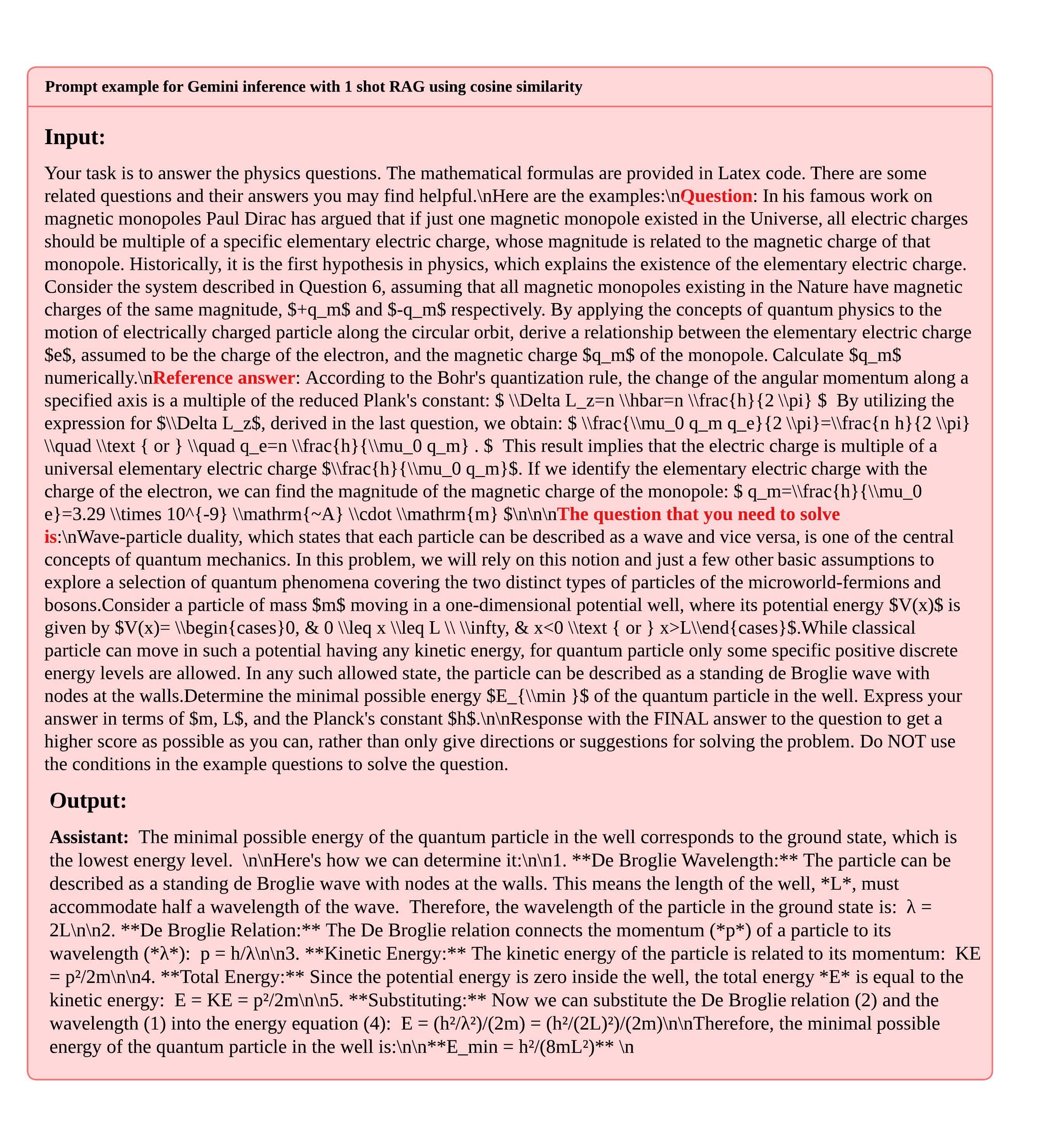} 
    \caption{An example for Gemini inference with 1-shot RAG using cosine similarity.}
    \label{Gemini_Example}
\end{figure}

\begin{figure}[H]
    \centering
    \includegraphics[width=0.5\textwidth]{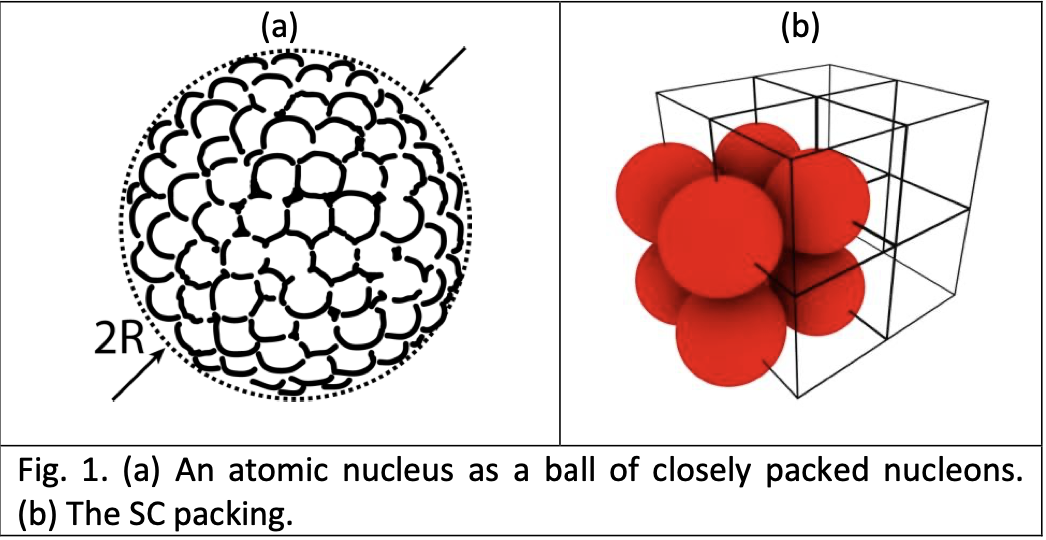} 
    \caption{The Image from the Question.}
    \label{fig:image1}
\end{figure}

Figure \ref{Gemini_Example} is an example of an answer provided by Gemini.

\label{Gemini_output}

\clearpage
\section{Multimodality Example}
\label{sec:Multimodality}
\textbf{Questions}: In a simple model, an atomic nucleus can be thought of as a ball consisting of closely packed nucleons [see Figure~1(a)], where the nucleons are hard balls of radius $r_N=0.85 \mathrm{fm}\left(1 \mathrm{fm}=10^{-15} \mathrm{~m}\right)$. The nuclear force is present only for two nucleons in contact. The volume of the nucleus $V$ is larger than the volume of all nucleons $A V_N$, where $V_N=\frac{4}{3} r_N^3 \pi$. The ratio $f=A V_N / V$ is called the packing factor and gives the percentage of space filled by the nuclear matter. Calculate what would be the packing factor $f$ if nucleons were arranged in a "simple cubic" (SC) crystal system, where each nucleon is centered on a lattice point of an infinite cubic lattice [see Figure~1(b)]. (0.3 points)

\textbf{Standard Answer:}
In the SC-system, in each of 8 corners of a given cube there is one unit (atom, nucleon, etc.), but it is shared by 8 neighboring cubes - this gives a total of one nucleon per cube. If nucleons are touching, as we assume in our simplified model, then \textcolor{red}{$a=2 r_N$} is the cube edge length a. The volume of one nucleon is then \textcolor{red}{$ V_N=\frac{4}{3} r_N^3 \pi=\frac{4}{3}\left(\frac{a}{2}\right)^3 \pi=\frac{4 a^3}{3 \cdot 8} \pi=\frac{\pi}{6} a^3 $} from which we obtain $ f=\frac{V_N}{a^3}=\frac{\pi}{6} \approx \textcolor{blue}{0.52} $

\textbf{GPT-4-Vision Output:(Score: 10)} In the "simple cubic" (SC) crystal system, each nucleon is centered on a lattice point, and because we're assuming the nucleons to be hard spheres, the sides of the cubic lattice cell must be twice the radius of a nucleon to accommodate one nucleon inside without overlapping with others.  Given that the radius of a nucleon \( r_N \) is 0.85 fm (fermi), which is \( 0.85 \times 10^{-15} \) meters, the side length \( a \) of the cubic cell will be:\textcolor{red}{$a = 2 \times r_N$} The volume \( V_N \) of a single nucleon (since it's a sphere) is:\textcolor{red}{$V_N = \frac{4}{3} \pi r_N^3$} Now, the volume $V_{SC}$ of the cubic cell which contains one nucleon would be: $V_{SC} = a^3 = (2r_N)^3 = 8r_N^3$ The packing factor \( f \) will be the ratio of the nucleon's volume to the volume of the cubic cell: $f = \frac{A V_N}{V} = \frac{V_N}{V_{SC}} $ Since each cell volume \( V_{SC} \) contains one nucleon for the simple cubic packing, we don't need to multiply \( V_N \) by \( A \), because \( A \) will cancel when we take the ratio for one cubic cell. Plugging in the expressions for \( V_N \) and \( V_{SC} \) gives: $f = \frac{\frac{4}{3} \pi r_N^3}{8 r_N^3} $ alculating this gives:$f = \frac{\pi}{6}$ Numerically, this evaluates to approximately:$f \approx \frac{3.14159}{6} \approx 0.5236$Or, in percentage terms:$f \approx 52.36\%$his means the packing factor or the percentage of space filled by the nuclear matter in a simple cubic arrangement is approximately \textcolor{blue}{52.36}\%.

\textbf{Gemini-Pro-Vision Output:(Score: 2)} I see the image you sent and can use it to understand the setup of the Kelvin water dropper.
The absolute value of the charge, $Q_0$, of each droplet as they separate from the tubes and at the instant the capacitor's charge is $q$ can be expressed as:
$Q_0=C q /\left(2 r_{\max }\right)$
where: $C$ is the capacitance. $q$ is the charge on the capacitor. $r_{\max }$ is the maximum radius of the droplet as it breaks free from the nozzle
This expression is derived based on the following:
1. Charge Quantization: Each droplet acquires a quantized charge, meaning its charge is a multiple of the fundamental electron charge (e). As the first droplet falls, it acquires a small charge imbalance, which is then shared equally between the two electrodes due to the capacitor.
4. Maximum Radius: The maximum radius, $r_{\max }$, relates to the amount of charge a droplet can hold due to its surface tension. A larger droplet can hold more charge.
Therefore, the absolute value of the charge on each droplet depends on the capacitance, the capacitor's charge, and the maximum radius the droplet can achieve before breaking free.
It's important to note that this expression neglects the effect mentioned in Part A-iii, where the charge on the droplet can induce a charge on the metal nozzle. This additional effect can slightly modify the charge acquired by the droplet.
I hope this helps! Let me know if you have any other questions.

\section{Runtime Analysis}

\label{app:runtime_details}

\begin{table}[ht]
\centering
\begin{tabular}{@{}lcc@{}}
\toprule
Algorithm & Pure Retrieval (s) & Retrieval + Embedding (s) \\ \midrule
Contriver & 23 & 255 \\
Dragon+ & 33 & 155 \\
Cosine similarity & 126 & 275 \\
BM25 & 7 & 7.4 \\ \bottomrule
\end{tabular}
\caption{Performance Comparison of Different RAG Algorithms in Second.}
\label{tab:rag_performance}

\end{table}

\clearpage
\section{Examples of answer errors}
\label{app:error_analysis}
\subsection{Error Type 1: Give guidelines instead of direct answer}

\paragraph{Problem} A ball is launched at speed $v$ at angle $\theta$ above the horizontal toward a vertical wall a distance$L$ away. It bounces elastically off the wall and falls back to its launch point. What was its initialspeed?

\paragraph{Reference Answer} $\frac{\sqrt{2Lg}}{\sqrt{\sin(2\theta)}}$

\paragraph{GPT-3.5's response} To solve this problem, we can consider the motion in two parts: the motion from the launch point to the wall, and then the motion from the wall back to the launch point.For the first part of the motion (from the launch point to the wall), we can consider the horizontal motion and the vertical motion separately. In the horizontal direction, the ball travels a distance $L$ with a constant horizontal velocity. In the vertical direction, the ball travels a vertical distance $h$, given by: $h = \frac{1}{2} g t^2$ where $t$ is the time taken to reach the wall, and $g$ is the acceleration due to gravity.Since the ball lands back at the launch point, the total time taken for the entire motion is $2t$. The vertical speed of the ball when it reaches the wall can be calculated as follows: $v_y = g t$. From the kinematic equation, we know that the vertical component of velocity when hitting the wall is equal to the initial vertical component of velocity multiplied by $-1$ (due to reflection of velocity), thus: $v \sin(\theta) = -g t$. Solving for $t$ gives: $t = \frac{v \sin(\theta)}{g}$.Since an elastic collision preserves kinetic energy, the speed of the ball upon reflection is the same as the incoming speed vector. Therefore, the total horizontal distance traveled by the ball is $2L$. The total horizontal distance can be calculated as follows: $2L = v \cos(\theta) \cdot 2t$. Substitute the expression for $t$: $2L = v \cos(\theta) \cdot \frac{2v \sin(\theta)}{g}$. Solving for $v$ gives: $v = \sqrt{\frac{g L}{\sin(2\theta)}}$.Therefore, the initial speed of the ball is $\sqrt{\frac{g L}{\sin(2\theta)}}$. \textbf{Score: 4}

\paragraph{Example 1 reterieved by cosine similarity}
A projectile is launched across flat ground at an angle $\theta$ to the horizontal and travels in the absence of air resistance. It rises to a maximum height $H$ and lands a horizontal distance $R$ away. What is the ratio $H / R$ ? Reference answer: $\frac{1}{4} \tan \theta$.

\paragraph{Example 2 reterieved by cosine similarity}
A cannonball is launched with initial velocity of magnitude $v_0$ over a horizontal surface. At what minimum angle $\theta_{\min}$ above the horizontal should the cannonball be launched so that it rises to a height $H$ which is larger than the horizontal distance $R$ that it will travel when it returns to the ground? Reference answer: $\theta_{\min} = 76^{\circ}$.

\paragraph{GPT-3.5's response using two-shot Retrieval-Augmented Generation}

We can break down the motion of the ball into horizontal and vertical components. When the ball hits the wall and bounces back, the vertical component of its velocity remains the same while the horizontal component changes direction. Let $v_x$ be the horizontal component of the initial velocity, and $v_y$ be the vertical component of the initial velocity. Since the motion is symmetrical, the time taken to return to the launch point is twice the time taken to reach the wall:$
t_{\text{total}} = 2 \cdot t_{\text{to wall}}
$The time taken to reach the wall can be found using the horizontal distance $L$ and the $x$-component of the initial velocity:$
L = v_x \cdot t_{\text{to wall}}
$From this, we solve for $t_{\text{to wall}}$:$
t_{\text{to wall}} = \frac{L}{v_x}
$Now, considering the vertical motion, at the maximum height $H$ the vertical component of the velocity is zero, hence:$
v_y^2 - 2 \cdot g \cdot H = 0
$where $g$ is the acceleration due to gravity. As the initial velocity is the magnitude of the total velocity:$
v = \sqrt{v_x^2 + v_y^2}
$From the above two equations we can solve for $v$. \textbf{Score: 3}

\paragraph{Analysis} In this case, the LLM only provided advisory comments, stating `From the above two equations we can solve for $v$'. without continuing to answer the question, resulting in a loss of some points.

\clearpage
\subsection{Error Type 2: Misguided by the retrieved examples.}

\paragraph{Problem} The change in the gas temperature due to the adiabatic expansion and contraction, as a result of the sound wave, takes the form:$T(x, t)=T_0-T_1(x) \cos (\omega t)$.Compute the amplitude $T_1(x)$ of the temperature oscillations in terms of $T_0, \gamma$, $a, k$ and $x$.

\paragraph{Reference Answer} Given:
- Temperature perturbation due to the sound wave: $T(x, t) = T_0 - T_1(x) \cos(\omega t)$,
- Adiabatic constant: $\gamma$,
- Displacement amplitude: $a$,
- Wave parameters: $k$ and $\omega$.
We know that for adiabatic processes, $pV^\gamma = \text{const}$, and for ideal gases, $pV = nRT$ where $n$ is the number of moles and $R$ is the gas constant.
From the ideal gas law, we have $p = \frac{nRT}{V}$. Let's consider the equilibrium state where $p = p_0$, $V = V_0$, and $T = T_0$.
Assuming small temperature perturbation from the equilibrium state:
$T = T_0 + T_1(x) \cos(\omega t)$.
Substitute into the ideal gas law:
$\frac{nR(T_0+T_1(x)\cos(\omega t))}{V_0} = p_0$.
Expanding this expression to first order in the perturbations, we get:
$T_1(x) = -\gamma T_0 a \sin(kx)$.
Therefore, the amplitude $T_1(x)$ of the temperature oscillations is $ -\gamma T_0 a \sin(kx)$ in terms of $T_0$, $\gamma$, $a$, $k$, and $x$.

\paragraph{GPT-3.5's response (Score: 6)} In a thermoacoustic system where the gas oscillates due to a standing sound wave, and there is weak thermal interaction between the gas and the tube, let us analyze the temperature changes at specific points on the tube. At point A, located at the edge of the tube, the gas parcel experiences maximum compression and minimum volume. As the gas undergoes an adiabatic expansion, it cools down, leading to a decrease in the tube's temperature at this point over time. At point B, the central part of the tube, the gas parcel is at its equilibrium position, neither at maximum compression nor expansion. Here, the temperature tends to remain constant as the gas parcel neither gains nor loses significant heat to the tube. At point C, another edge of the tube, the gas parcel is at maximum expansion and volume. Upon adiabatic compression, the gas heats up, resulting in an increase in the tube's temperature at this point. Therefore, in this scenario of weak thermal interaction, over a prolonged period, the temperature at point A decreases, remains unchanged at point B, and increases at point C. 

\paragraph{Example 1 retrieved by cosine similarity}
Consider a longitudinal wave traveling along the $x$ direction in the rod. For a cross section at $x$ when the rod is unstressed, let $\xi(x, t)$ be its displacement at time $t$ and assume $ \xi(x, t)=\xi_0 \sin k(x-u t) $ where $\xi_0$ and $k$ are constants. Determine the corresponding velocity $v(x, t)$, strain $S(x, t)$, and pressure $p(x, t)$ as a function of $x$ and $t$. Reference answer: The velocity $v$ is related to the displacement $\xi$ as in a simple harmonic motion (or a uniform circular motion, as shown in Figure A2) of angular frequency $\omega=k u$. Therefore, if $\xi(x, t)=\xi_0 \sin k(x-u t)$, then $ v(x, t)=-k u \xi_0 \cos k(x-u t) . $  The strain and pressure are related to velocity as in Problem (a). Hence, $  S(x, t)  =-v(x, t) / u=k \xi_0 \cos k(x-u t) p(x, t)  =\rho u v(x, t)=-k \rho u^2 \xi_0 \cos k(x-u t) =-Y S(x, t)=-k Y \xi_0 \cos k(x-u t)$.

\paragraph{Example 2 retrieved by cosine similarity}
In the uniform background magnetic field, theplasma density is nonuniform in $x$. For simplicity, we assume that the temperature and the distribution of the ions and electrons are the same. Thus, the plasma pressure can be expressed as $p(x)=k T\left[n_i(x)+n_e(x)\right]=2 k T n(x)=2 k T\left(n_0+\alpha x\right)$, Where $B, T, k, n_0$, and $\alpha$ are positive constants, $n_i(x)$ and $n_e(x)$ are the number densities of the ions and electrons. The diamagnetic field is given by $B_{m x}=\mu_o M$. Please explain the generation mechanism of the electric current by a schematic drawing. Reference answer: Since the magnetic field and the plasma are uniform $z$, the orbits of ions and electrons can project into in the $x-y$ plane. From the results of sub\_question\_number: 1, we know that an ion has a left-hand circular motion and an electron has a right-hand circular motion. Due to the linear increase of the plasma density in $x$, the number of ions with upward motion is less than that with downward motion at a given $x$ position, which leads a net upward ion flow. Similarly, electrons have a net downward flow. Combining the ion and election flows, we have a net upward electric current as illustrated below in schematic drawing.

\paragraph{GPT-3.5's response using two-shot Retrieval-Augmented Generation (Score: 4)}
Given:
- Temperature perturbation due to the sound wave: $T(x, t) = T_0 - T_1(x) \cos(\omega t)$,
- Adiabatic constant: $\gamma$,
- Displacement amplitude: $a$,
- Wave parameters: $k$ and $\omega$.
We know that for adiabatic processes, $pV^\gamma = \text{const}$, and for ideal gases, $pV = nRT$ where $n$ is the number of moles and $R$ is the gas constant.
From the ideal gas law, we have $p = \frac{nRT}{V}$. Let's consider the equilibrium state where $p = p_0$, $V = V_0$, and $T = T_0$.
Assuming small temperature perturbation from the equilibrium state:
$T = T_0 + T_1(x) \cos(\omega t)$.
Substitute into the ideal gas law:
$\frac{nR(T_0+T_1(x)\cos(\omega t))}{V_0} = p_0$.
Expanding this expression to first order in the perturbations, we get:
$T_1(x) = -\gamma T_0 a \sin(kx)$.
Therefore, the amplitude $T_1(x)$ of the temperature oscillations is $ -\gamma T_0 a \sin(kx)$ in terms of $T_0$, $\gamma$, $a$, $k$, and $x$.

\paragraph{Analysis} The original question was relatively short, but the retrieved questions are quite lengthy, inevitably focusing more attention on the retrieved content. After the RAG process, it is evident that the LLM listed conditions that did not exist in the original question.

\clearpage
\subsection{Error Type 3: Wrong Retrieval.}

\paragraph{Problem} his part deals with the usage of microwave radiation (radiated by the magnetron antenna into the food chamber) for cooking, that is, heating up a lossy dielectric material such as water, either pure or salty (which is our model for, say, soup). An electric dipole is a configuration of two equal and opposite electric charges $q$ and $-q$ a small distance $d$ apart. The electric dipole vector points from the negative to the positive charge, and its magnitude is $p=q d$. A time-dependent electric field $\vec{E}(t)=E(t) \hat{x}$ is applied on a single dipole of moment $\vec{p}(t)$ with constant magnitude $p_0=|\vec{p}(t)|$. The angle between the dipole and the electric field is $\theta(t)$.Write expressions for both the magnitude of the torque $\tau(t)$ applied by the electric field on the dipole and the power $H_i(t)$ delivered by the field to the dipole, in terms of $p_0, E(t), \theta(t)$ and their derivatives.

\paragraph{Reference Answer}The torque at time $t$ is given by $\tau(t)=-q d \sin [\theta(t)] E(t)=-p_0 \sin [\theta(t)] E(t)$,hence the instantaneous power delivered to the dipole by the electric field is $H_i(t)=\tau(t) \dot{\theta}(t)=-p_0 E(t) \sin \theta(t) \dot{\theta}(t)=E(t) \frac{d}{d t}\left(p_0 \cos \theta(t)\right)=E(t) \frac{d p_x(t)}{d t}$.

\paragraph{GPT-3.5's response (Score: 9)} The torque $\tau(t)$ applied by the electric field on the dipole can be expressed as the cross product of the electric dipole moment $\vec{p}(t)$ and the electric field $\vec{E}(t)$:\[\boldsymbol{\tau}(t) = \vec{p}(t) \times \vec{E}(t)\]Expanding the cross product using vector algebra gives:\[\boldsymbol{\tau}(t) = p_0 \hat{p}(t) \times E(t) \hat{x} = p_0 E(t) \sin(\theta(t)) \hat{z}\]where $\hat{p}(t)$ is the unit vector pointing along the direction of the dipole moment $\vec{p}(t)$ and $\hat{z}$ is the unit vector along the $z$-axis.The power $P(t)$ delivered by the field to the dipole can be calculated using the dot product of the torque $\boldsymbol{\tau}(t)$ and the angular velocity $\boldsymbol{\omega}(t)$ of the dipole:\[P(t) = \vec{\tau}(t) \cdot \vec{\omega}(t)\]As the electric field $\vec{E}(t)$ is time-dependent and not directly causing the dipole to rotate, the power delivered to the dipole is due to changes in the potential energy of the system, which can be related to the time derivative of the dipole moment $\vec{p}(t)$. Therefore, the power $H_i(t)$ delivered by the field to the dipole can be expressed as:\[H_i(t) = \frac{d}{dt} (\vec{p}(t) \cdot \vec{E}(t))\]By substituting the given values, we obtain the expression for the power:\[H_i(t) = \frac{d}{dt} (p_0 E(t) \cos(\theta(t)))\]  

\paragraph{Example 1 retrieved by cosine similarity}
In the figure below, a simplified scheme of a mass-spectrometer is given. It is a device for measuring the masses of molecules. The substance under investigation is ionised by heating up to a temperature $T$ on a hot filament (molecules undergo a single-electron ionisation). The ions are accelerated using voltage $U$. At first, let us neglect the thermal energy of the ions $(e U \gg k T$, where $e$ is elementary charge and $k$ - the Boltzmann's constant). A narrow beam of accelerated ions enters a region with magnetic field. For the sake of simplicity, let us assume that the region has a rectangular shape, and the magnetic field is homogeneous inside it. The magnetic fields deflects the ions and depending on their mass, they may hit the detector. Let us assume that those ions which hit the centre of the detector enter and exit the region with magnetic field perpendicularly with its boundary, and the distance between the entry and exit points is l (see figure). 1) Express the mass $M$ of those ions which hit the centre of the detector via the quantities $B$ $l, U$ and $e$.Reference answer:The trajectory of a charged particle in the magnetic field is circle of radius $R=l / \sqrt{2}$. Lorenz force is responsible for the acceleration, $\mathrm{Bev}=M v^{2} / R$ hence $B e R=p$. Substituting $p^{2}=2 M U e=$ $B^{2} e^{2} R^{2}$, we obtain $M=B^{2} l^{2} e / 4 U$.

\paragraph{Example 2 retrieved by cosine similarity}
One way of measuring the atmospheric electric field, and hence $\sigma_0$, is with the system shown in the diagram. A pair of metal quadrants, insulated from ground but connected to each other, are mounted just underneath a grounded uniformly rotating disk with two quadrant-shaped holes cut in it. (In the diagram, the spacing has been exaggerated in order to show the arrangement.) Twice in each revolution the insulated quadrants are completely exposed to the field, and then (1/4 of a period later) are completely shielded from it. Let $T$ be the period of revolution, and let the inner and outer radii of the insulated quadrants be $r_1$ and $r_2$ as shown. Take $t=0$ to be an instant when the insulated quadrants are completely shielded. Obtain expressions that give the total charge $q(t)$ induced on the upper surface of the insulated quadrants as a function of time between $t=0$ and $t=T / 2$, and sketch a graph of this variation. [The effects of the atmospheric ion current can be ignored in this situation.]

\paragraph{GPT-3.5's response using two-shot Retrieval-Augmented Generation (Score: 4)}
The torque applied by an electric field on a dipole is given by $\boldsymbol{\tau}(t) = \vec{p}(t) \times \vec{E}(t)$.
Since $\vec{p}(t) = p_0 \cos(\theta(t)) \hat{y}$ and $\vec{E}(t) = E(t) \hat{x}$, the torque would be:
\[\boldsymbol{\tau}(t) = p_0 E(t) \cos(\theta(t)) \hat{z}\]
The power $P$ delivered by the electric field to the dipole is given by $\frac{dW}{dt} = \vec{F} \cdot \vec{v}$, where $\vec{F}$ is the force on the dipole and $\vec{v}$ is its velocity. Since power is force times velocity, we get: 
\[P = \vec{F} \cdot \vec{v} = \boldsymbol{\tau} \cdot \vec{\omega}\]
Where $\vec{\omega}$ is the angular velocity of the dipole. 
Substituting $\boldsymbol{\tau}(t) = p_0 E(t) \cos(\theta(t)) \hat{z}$ and $\vec{\omega}(t) = -\dot{\theta}(t) \hat{z}$, we find:
\[H_i(t) = \boldsymbol{\tau}(t) \cdot \vec{\omega}(t) = -p_0 E(t) \dot{\theta}(t) \cos(\theta(t))\].

\paragraph{Analysis} Although the original topic is about wave radiation, and the searched topics are also related to waves, the specific content and background of the topics are different: Example 1 describes a simplified model of a mass spectrometer. Example 2 discusses a method for measuring the electric field of the atmosphere.

\section{Additional Results}
\label{app:finegain_results}

\subsection{Experimental results of problems with different difficulty levels.}
We provide the number of questions of different difficulty levels, as shown in Table~\ref{tab:difficulty_distribution}. The number of questions whose answer scores are improved from zero to non-zero after RAG are shown in Table~\ref{tab:model_improvement}. 
Average score of LLMs with RAG over different difficulty level problems are provided in Table~\ref{scores}.

\begin{table}[ht]
\centering
\begin{tabular}{@{}lcccc@{}}
\toprule
Category     & Easy & Medium & Hard & Superhard \\ 
\midrule
Count        & 51   & 74     & 148  & 117       \\ 
\bottomrule
\end{tabular}
\caption{Distribution of Question Difficulty.}
\label{tab:difficulty_distribution}
\end{table}

\begin{table}[ht]
\centering
\begin{tabular}{@{}lccccc@{}}
\toprule
Model & Method & Easy & Medium & Hard & Superhard \\ 
\midrule
GPT-3.5 & Base & 9.41 & 5.89 & 3.76 & 1.15 \\
GPT-3.5 & BM25 & 6.51 & 4.74 & 3.55 & 2.45 \\
GPT-3.5 & Dragon+ & 6.63 & 4.08 & 3.68 & 2.38 \\
GPT-3.5 & Contriver & 6.35 & 4.85 & 3.72 & 2.48 \\
GPT-3.5 & Cos & 6.57 & 5.04 & 3.75 & 2.62 \\ \hline
GPT-4 & Base & 8.45 & 7.15 & 6.14 & 4.91 \\
GPT-4 & BM25 & 8.51 & 6.81 & 5.78 & 4.42 \\
GPT-4 & Dragon+ & 8.53 & 7.07 & 5.60 & 4.38 \\
GPT-4 & Contriver & 8.94 & 6.70 & 5.29 & 4.60 \\
GPT-4 & Cos & 8.45 & 6.45 & 5.41 & 4.17 \\
\bottomrule
\end{tabular}
\caption{Average score of GPT-3.5 and GPT4 with RAG in different difficulty level problems.}
\label{scores}
\end{table}

\begin{table}[ht]
\centering
\begin{tabular}{@{}lccccc@{}}
\toprule
Model & BM25 & Cos & Dragon+ & Contriver & Average \\ 
\midrule
GPT-4 & 6 & 7 & 5 & 9 & 6.75\\
GPT-3.5 & 18 & 18 & 24 & 19 & 19.75\\
Gemini & 26 & 22 & 23 & 22 & 23.25\\
DeepSeek-Math & 10 & 11 & 12 & 12 & 11.25\\
\bottomrule
\end{tabular}
\caption{Number of questions whose answer scores are raised from zero to non-zero after RAG.}
\label{tab:model_improvement}
\end{table}
\label{sec:appendix}

\end{document}